%% file: main.tex
\documentclass[conference]{IEEEtran}
\IEEEoverridecommandlockouts
\usepackage{cite}
\usepackage{amsmath,amssymb,amsfonts}
\usepackage{algorithmic}
\usepackage{graphicx}
\usepackage{textcomp}
\usepackage{xcolor}
\def\BibTeX{{\rm B\kern-.05em{\sc i\kern-.025em b}\kern-.08em
    T\kern-.1667em\lower.7ex\hbox{E}\kern-.125emX}}
\usepackage{cite}
\usepackage{amsmath,amssymb,amsfonts}

\usepackage{graphicx}
\usepackage{textcomp}

\usepackage{booktabs}
\usepackage{makecell}
\usepackage{multicol, multirow}
\usepackage[caption=false,font=normalsize]{subfig}
\usepackage{xcolor, xspace}
\usepackage{algorithm}
\usepackage{algorithmic}
\usepackage{nccmath}
\usepackage{threeparttable}

\newcommand{\etal}{\emph{et~al.}\xspace}
\newcommand{\eg}{\emph{e.g.},\xspace}
\newcommand{\ie}{\emph{i.e.},\xspace}
\newcommand{\etc}{etc.\xspace}

\newcommand\figref[1]{Figure~\ref{#1}}

\newcommand\equref[1]{Equ.~(\ref{#1})}
\newcommand\tabref[1]{Table~\ref{#1}}
\newcommand\secref[1]{Sec.~\ref{#1}}

\newcommand{\sysname}{{HTSF}\xspace}

\begin{document}

\title{Combating Distribution Shift for Accurate Time Series Forecasting via Hypernetworks}

\author{\IEEEauthorblockN{1\textsuperscript{st} Wenying Duan}
\IEEEauthorblockA{\textit{School of Mathematics and Computer Sciences} \\
\textit{Nanchang University}\\
Nanchang, China \\
wenyingduan@ncu.edu.cn}
\and
\IEEEauthorblockN{2\textsuperscript{nd} Xiaoxi He}
\IEEEauthorblockA{\textit{ Computer Engineering and Networks
Laboratory} \\
\textit{ETH Zurich}\\
Zurich, Switzerland \\
hex@ethz.ch}
\and
\IEEEauthorblockN{3\textsuperscript{rd} Lu Zhou}
\IEEEauthorblockA{\textit{Department of Civil and Environmental Engineering} \\
\textit{The Hong Kong Polytechnic University}\\
HongKong\\
lu.lz.zhou@polyu.edu.hk}
\and
\IEEEauthorblockN{4\textsuperscript{th} Lothar Thiele}
\IEEEauthorblockA{\textit{ Computer Engineering and Networks
Laboratory} \\
\textit{Zurich, Switzerland}\\
Zurich, Switzerland \\
 thiele@ethz.ch}
\and
\IEEEauthorblockN{5\textsuperscript{th} Hong Rao$^{*}$}
\IEEEauthorblockA{\textit{school of software} \\
\textit{Nanchang University}\\
Nanchang, China \\
raohong@ncu.edu.cn}
}

\maketitle

\begin{abstract}
Time series forecasting has widespread applications in urban life ranging from air quality monitoring to traffic analysis.
However, accurate time series forecasting is challenging because real-world time series suffer from the distribution shift problem, where their statistical properties change over time. 
Despite extensive solutions to distribution shifts in domain adaptation or generalization, they fail to function effectively in unknown, constantly-changing distribution shifts, which are common in time series. 
In this paper, we propose Hyper Time-Series Forecasting (\sysname), a hypernetwork-based framework for accurate time series forecasting under distribution shift. 
\sysname jointly learns the time-varying distributions and the corresponding forecasting models in an end-to-end fashion.  
Specifically, \sysname exploits the hyper layers to learn the best characterization of the distribution shifts, generating the model parameters for the main layers to make accurate predictions. 
We implement \sysname as an extensible framework that can incorporate diverse time series forecasting models such as RNNs and Transformers.  
Extensive experiments on 9 benchmarks demonstrate that \sysname achieves state-of-the-art performances.
\end{abstract}


\input{body/introduction}
\input{body/related}

\input{body/preliminary}

\input{body/method}

\input{body/experiment}

\input{body/analysis}

\input{body/conclusion}



\clearpage

\bibliographystyle{IEEEtran.bst}
\bibliography{IEEEfull,cites}
\end{document}

%% file: body/introduction.tex
\section{Introduction}
\label{sec:intro}
\setcounter{footnote}{0}
\renewcommand*{\thefootnote}{\fnsymbol{footnote}}
\footnotetext[1]{Corresponding Author}
Time-series forecasting is crucial for various data analytics domains, including air quality monitoring \cite{zhang2017cautionary}, renewable energy production \cite{en12020215}, human activity recognition \cite{almaslukh2017effective}, electricity consumption planning \cite{zhou2021informer}, urban traffic analysis~\cite{zhou2021informer} \etc
The past years have witnessed diverse time-series forecasting models ranging from the conventional statistical approaches \eg auto-regressive integrated moving average (ARIMA) \cite{hamilton2020time} to the more recent deep learning based models such as recurrent neural networks (RNNs) \cite{connor1992recurrent}, transformers, \cite{vaswani2017attention} and their variants \cite{lai2018modeling, le2020probabilistic, salinas2020deepar, zhou2021informer}.
As deep learning based models make few assumptions on the temporal structures of time series, they are preferable in modeling complex, long-sequence time series. They have demonstrated a notable improvement in prediction accuracy than statistical models \cite{lai2018modeling}. 

\begin{figure}[t]
\centering
\includegraphics[width=0.98\linewidth]{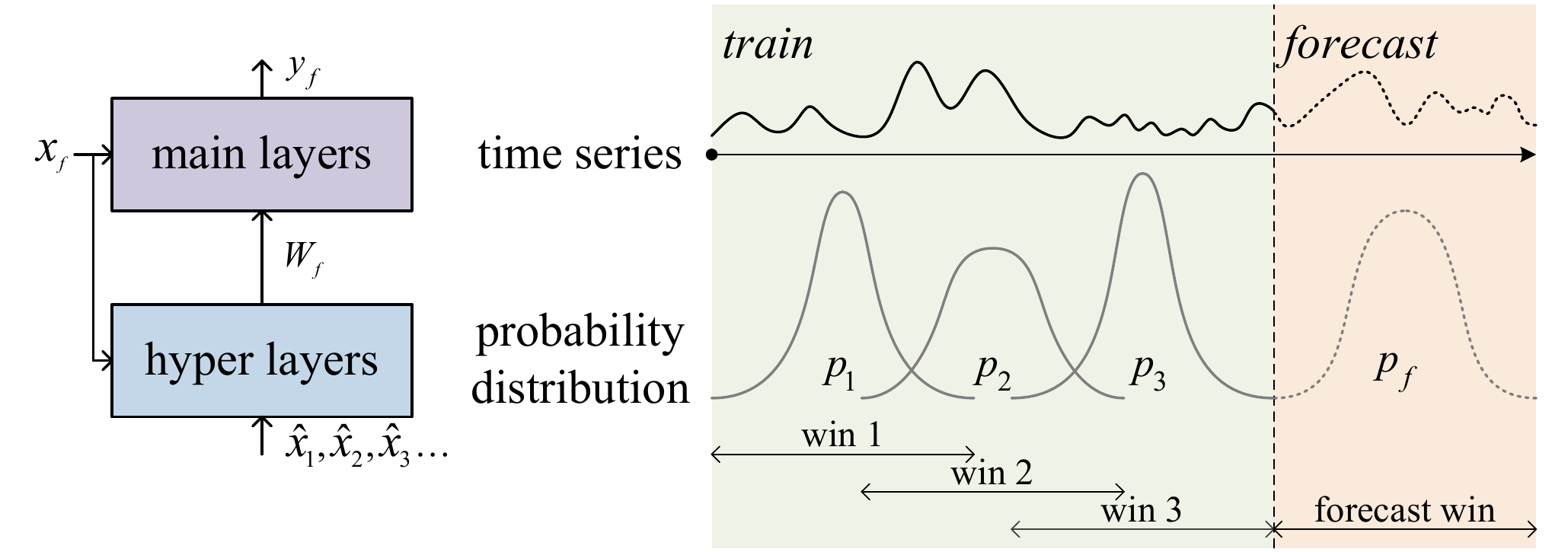}
\caption{Illustration of distribution shifts in time series forecasting and our solution. Given sliding windows win 1, 2, 3 and so on, the data distributions $p_{1}$, $p_{2}$, $p_{3}$, $\ldots$ vary across windows, and they are likely to differ from the distribution $p_{f}$ in the forecasting window. Our solution is to explicitly model and learn the data distributions from the historical sequences $\hat{\mathbf{x}}_{1}$, $\hat{\mathbf{x}}_{2}$, $\hat{\mathbf{x}}_{3}$, $\ldots$ via hyper layers, which generate the model parameters $W_{f}$ for the main layers. The main layers then take $\mathbf{x}_{f}$ as input and generate the predictions $y_{f}$.}
\label{fig:intro}
\end{figure}

A unique challenge in time series forecasting is that they are often non-stationary, resulting in temporal shifts in their distributions. 
This distribution shift phenomenon causes discrepancies between the distributions of the training and the testing data, which degrades the performance of forecasting models \cite{DBLP:conf/alt/KuznetsovM14, du2021adarnn}.
Such distribution shift also exists among the input sequences of the training data, which makes it challenging to train a model that generalizes well to unknown data \cite{DBLP:conf/cvpr/TorralbaE11, DBLP:conf/cvpr/TzengHSD17}.
\figref{fig:intro} illustrates the distribution shift problem in time series forecasting.
Consider three sliding windows 1, 2, and 3 to segment the time series for training. 
The probability distributions of the time series $p_{1}$, $p_{2}$, $p_{3}$ may vary across windows, and they are likely to differ from the distribution  $p_{f}$ in the unseen forecast window.
That is, $p_{1} \neq p_{2} \neq p_{3} \neq p_{f}$, which makes accurate forecasting for window $p_{f}$ challenging.

Distribution shift is usually tackled by domain generalization \cite{DBLP:conf/ijcai/0001LLOQ21} or domain adaptation \cite{DBLP:conf/cvpr/TzengHSD17, wang2018visual}. 
The idea is to learn the common knowledge transferable between domains despite the differences in their distributions \cite{DBLP:conf/ijcai/0001LLOQ21}. 
However, it is non-trivial to apply these techniques to time series \cite{kim2021reversible, du2021adarnn}.
This is because the distributions in time series change constantly (\eg $p_{1}$, $p_{2}$, $p_{3}$ in \figref{fig:intro}), and it is unknown \textit{how to best characterize the distributions} upon which to learn the common knowledge \cite{du2021adarnn}. 
The state-of-the-art, ADARNN \cite{du2021adarnn} proposes to characterize the \textit{worst-case} distribution shift in time series leveraging the principle of maximum entropy.
Nevertheless, this approach is sub-optimal because \textit{(i)} the training data may not always contain the worst-case distribution shift; and \textit{(ii)} it under-utilizes the diversity of the distribution shifts in the time series. 


In this paper, we propose a hypernetwork-based framework called Hyper Time-Series Forecasting (\sysname) to mitigate the distribution shift problem for accurate time series forecasting.
The idea is to jointly learn the common knowledge and the corresponding distributions with an end-to-end hypernetwork architecture.
As shown in \figref{fig:intro}, a hypernetwork consists of hyper and main layers, where the hyper layers generate model parameters $W_f$ conditioned on the inputs $\mathbf{x}_{f}$ for the main layers to make predictions $y_{f}$.
We exploit the hyper layers to learn the best characterization of the distribution shifts and the main layers to learn the corresponding forecast model.
Our strategy is advantageous in the following aspects.
\textit{(i)} 
The end-to-end learning scheme eliminates the assumptions on the worst-case distribution shift in the training data and makes full use of the various distribution shifts in the training data.
\textit{(ii)} 
The hypernetwork architecture is capable of modeling complex feature representations in both the distributions and the common transferable knowledge, and it applies to various deep learning based time series models.


Our main contributions are summarized as follows:
\begin{itemize}
    \item 
    We propose \sysname, a hypernetwork-based framework for accurate time series forecasting. 
    To the best of our knowledge, it is the first study that addresses the distribution shift problem in time series via hypernetworks. 
    \item 
    \sysname offers a generic, extensible, and end-to-end solution to the distribution shift problem in time series that makes no assumptions on the training data and applies to diverse deep learning based time series models such as RNNs and transformers.
    \item 
    We evaluate \sysname on 9 time series forecasting benchmarks.
    Experimental results show that \sysname achieves state-of-the-art accuracy in both short-term and long-term time series forecasting. 
    It also yields comparable performance on spatiotemporal data to the state-of-the-art graph neural network based solutions \cite{DBLP:conf/nips/0001YL0020, DBLP:conf/aaai/LiZ21}.
\end{itemize}

In the rest of this paper, we review related work in \secref{sec:related}, present the problem statement in \secref{sec:problem} and introduce our method in \secref{sec:method}.
We report the performance in \secref{sec:exp} and ablation studies in \secref{sec:analysis}, and finally conclude in \secref{sec:conclusion}.

%% file: body/related.tex
\section{Related Work}
\label{sec:related}
Our work is related to the following categories of research.

\subsection{Time Series Forecasting}
For time series forecasting, we distinguish between statistic models and deep learning based models.
Statistic models\cite{1987Time, DBLP:conf/uksim/AriyoAA14} are interpretable and theoretically sound.
But they often suffer from heavy crafting on data pre-processing and labor-intensive feature engineering, which may fail to capture complex patterns in time series.
Recently, deep learning based models have attracted increasing attention due to their outstanding performance in time series analysis.
For example, RNN-based models \cite{en12020215, lai2018modeling, le2019shape} prove effective in capturing both short-term and long-term patterns. 
Transformer-based models \cite{li2019enhancing, zhou2021informer} have shown remarkable performance for extremely long sequences. 
However, they are not robust to the distribution shift problem and suffer from  performance degradation \cite{du2021adarnn, kim2021reversible}.

\subsection{Distribution Shift}
Domain adaptation \cite{DBLP:conf/cvpr/TzengHSD17, wang2018visual} and domain generalization \cite{DBLP:conf/ijcai/0001LLOQ21} are generic solutions to the distribution shift problem. 
Domain adaptation focuses on the disparity difference between the source and target domain, whereas domain generalization aims to learn a generalizable model from various source domains. 
However, it is non-trivial to apply these techniques to time series. 
For example, some research reports \cite{jiang2021regressive} that domain adversarial learning, a prevailing domain adaptation technique, may be inappropriate for regression, while time series forecasting is often a regression task. 
Furthermore, since the distributions in time series change constantly, it is not straightforward how to define multiple source domains in non-stationary time series \cite{du2021adarnn}. 

A few pioneer studies \cite{kim2021reversible, du2021adarnn} explore solving the distribution shift problem in time series forecasting.
RevIN \cite{kim2021reversible} proposes an instance-level normalization technique to reduce the discrepancy in data distributions such as mean and variance.
Our work is orthogonal to RevIN by adopting representation learning to enhance the generalization of forecasting models. 
Our work is most related to ADARNN \cite{du2021adarnn}, which first attempts to characterize the worst-case distribution shift in time series and then matches the distributions to learn a generalizable model.
We advance ADARNN by eliminating the worst-case assumption and fully exploiting the diversity of the distribution shifts in the training data.

\subsection{Hypernetworks}
A hypernetwork \cite{ha2016hypernetworks} utilizes a primary neural network $g$ to generate weights $\theta_{f}$ for a second network $f$. 
The network $f$ with the weights $\theta_{f}$ can then be applied for specific inference tasks. 
Hypernetworks have demonstrated superior performances on many benchmarks \cite{DBLP:conf/iclr/ZhangRU19, von2020continual}, due to their ability to adapt $f$ for different inputs, which allows the hypernetwork to model tasks more effectively \cite{galanti2020modularity}.
Hypernetworks have been applied in few-shot learning \cite{bertinetto2016learning}, continual learning \cite{von2020continual}, efficient parameter fine-tuning \cite{DBLP:conf/acl/MahabadiR0H20, duan2021injecting} \etc
A few studies have also introduced hypernetworks in time series forecasting such as traffic prediction \cite{DBLP:conf/kdd/PanLW00Z19}.
However, it does not explicitly account for the distribution shift problem, and thus results in sub-optimal accuracy.   
To the best of our knowledge, we are the first to adopt hypernetworks to tackle the distribution shift problem in time series forecasting.

%% file: body/preliminary.tex
\section{Problem Statement}
\label{sec:problem}
We consider a standard multi-variate time series forecasting problem as follows.
A time series $\mathcal{X}$ with the corresponding label sequence $\mathcal{Y}$, is segmented into a set of sequences with labels $\mathcal{D}=\left\{\mathbf{x}^{(i)}, \mathbf{y}^{(i)} \right\}_{i}^{N}$, where $\mathbf{x}^{(i)} \in \mathbb{R}^{d_{\mathbf{x}} \times T_{\mathbf{x}}}$ is a sub-sequence of $\mathcal{X}$, $\mathbf{y}^{(i)} \in \mathbb{R}^{d_{\mathbf{y}} \times T_{\mathbf{y}}}$ is the corresponding label form $\mathcal{Y}$. 
$N$, $d_{\mathbf{x}}$, $d_{\mathbf{y}}$, $T_{\mathbf{x}}$, and $T_{\mathbf{y}}$ denote the number of sequences, the dimension of input variables, the dimension of label variables, the input length and the prediction length, respectively. 
Our objective is to learn a model $\mathcal{M}: \mathrm{x}_{i} \rightarrow \mathrm{y}_{i}$ from $\mathcal{D}_{train}$ for unseen sequences in $\mathcal{D}_{test}$, where $\mathcal{D}_{train}$ and $\mathcal{D}_{test}$ are the training and test sets from $\mathcal{D}$.

We aim to learn model $\mathcal{M}$ in presence of \textit{distribution shift} in the time series, which is described by the following assumptions.
\begin{itemize}
    \item 
    There is discrepancy between the marginal probability distribution $P_{\mathcal{D}_{train}}$ of the training set and the distribution $\mathcal{D}_{test}$ of the test set, \ie $P_{\mathcal{D}_{train}}\neq P_{\mathcal{D}_{test}}$.
    Note that $P_{\mathcal{D}_{train}}(Y|X)=P_{\mathcal{D}_{test}}(Y|X)$ because the underlying laws governing the inputs and the outputs of the time series usually stay the same \cite{du2021adarnn}. 
    \item
    The marginal probability distribution of each $\mathbf{x}^{(i)}$ from $\mathcal{D}_{train}$ may differ, although their conditional distributions are the same.
    That is $P_{\mathcal{D}}(\mathbf{x}^{(i)}) \neq P_{\mathcal{D}}(\mathbf{x}^{(j)})$, $1 \leq i \neq j \leq N$, $P_{\mathcal{D}}\left(\mathbf{y}^{(i)} \mid \mathbf{x}^{(i)}\right)=P_{\mathcal{D}}\left(\mathbf{y}^{(j)} \mid \mathbf{x}^{(j)}\right)$.
\end{itemize}
The discrepancy between the training and test set highlights the necessity to account for distribution shift for accurate predictions, while that among the sub-sequences within the training set explains the challenges of effectively learning the model $\mathcal{M}$.




%% file: body/method.tex
\section{Method}
\label{sec:method}
This section presents \sysname, our hypernetwork-based framework for time series forecasting under distribution shift.
\sysname is a model-agnostic framework that can incorporate various deep learning based time series forecasting models such as RNNs and Transformers.
For ease of presentation, we explain \sysname using a gated recurrent unit (GRU) as an example forecasting model.

\subsection{\sysname Overview}
\label{sec:method:overview}
\begin{figure}[t]
  \centering
  \includegraphics[width=0.45\textwidth]{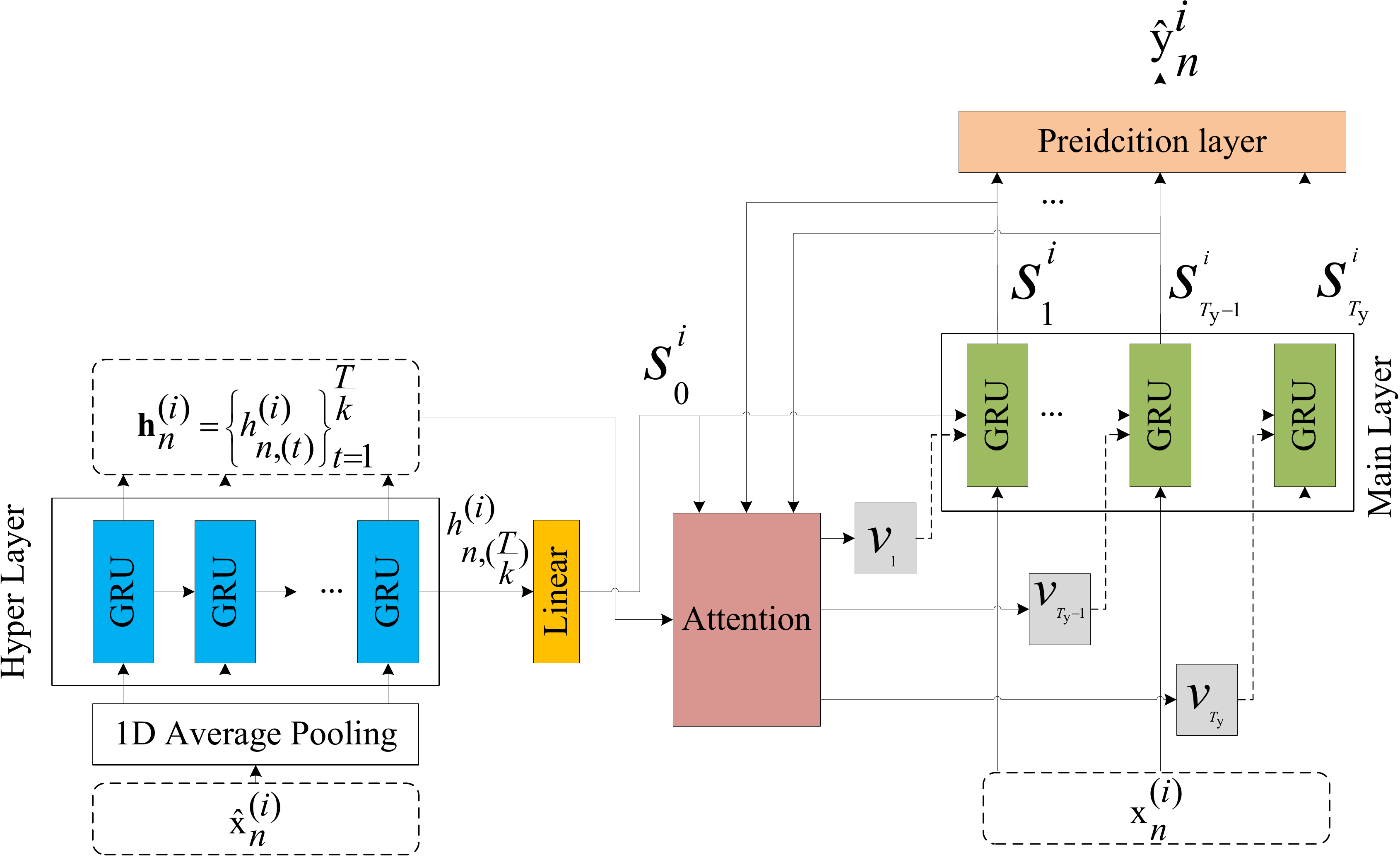} 
  \caption{An overview of \sysname, where we employ GRUs as the example forecasting model. The corresponding hypernetwork is termed HyperGRU.}
  \label{fig:overview}
\end{figure}

\figref{fig:overview} depicts the overview of \sysname with GRUs as the underlying forecasting model, which we call HyperGRU.
It mainly consists of two modules.
\begin{itemize}
    \item \textit{Hyper Layers}: 
    These layers learn to characterize the distributions from the historical sequences. 
    The learned distribution characterizations are then used to generate the weights of the main layers.
    We employ bidirectional GRU (BiGRU) for the hyper layers (see \secref{sec:method:meta}).
    \item \textit{Main Layers}: 
    These layers adapt their weights according to the learned distribution characterizations from the hyper layers as well as their own internal states through an attention mechanism.
    As a result, the main layers can leverage the common knowledge extracted from the diverse distribution shifts to make accurate predictions.
    The core part of the main layers is a simple GRU cell (see \secref{sec:method:main}).
\end{itemize}

The hyper and main layers of HyperGRU interact as follows.
Let $\widehat{\mathcal{X}}^{(i)}$ be the historical data of $\mathbf{x}^{(i)}$.
$S= \left\{\hat{\mathbf{x}}_{n}^{(i)}\right\}_{n=1}^{L}$ is the historical dataset generated by  $\widehat{\mathcal{X}}^{(i)}$ with a sliding window of a fixed length $T$, where $\hat{\mathbf{x}}_{n}^{(i)} \in \mathbb{R}^{d_{\mathbf{x}} \times T}$, $L=\left|\widehat{\mathcal{X}}^{(i)}\right|-T+1$.
The hyper layers take all $\hat{\mathbf{x}}_n^{(i)}$ from $S$ as inputs and output a set of representations of the historical marginal probability distributions.
These representations and input at the current time step from $\mathbf{x}^{(i)}$ are then combined through the attention mechanism to generate the weights for the main layers at the next time step. 
Finally, the main layers make predictions as conventional time series forecasting methods.

Note that our method makes full use of the various distribution shifts in historical data $S$ and aggregates the common transferable knowledge from $S$ to mitigate the distribution shift problem for accurate time series forecasting.
In comparison, prior proposals either ignore the distribution shift problem \cite{lai2018modeling, le2020probabilistic, zhou2021informer} or under-utilize the diverse distribution shifts in the time series \cite{du2021adarnn}.

\subsection{Hyper Layers}
\label{sec:method:meta}
The hyper layers are responsible to learn the complex feature representations of various distribution shifts.
We use BiGRU to collectively encode all sequences in the historical data $S= \left\{\hat{\mathbf{x}}_{n}^{(i)}\right\}_{n=1}^{L}$.
\begin{itemize}
    \item 
    We first apply an 1D average pooling with kernel size $k$ and stride $k$ over $\hat{\mathbf{x}}_{n}^{(i)}$ on time dimension to down sample $\hat{\mathbf{x}}_{n}^{(i)}$ into its $\frac{1}{k}$ slice: 
    \begin{equation}\label{eq:avgpool}
        \medop{\overline{\mathbf{x}}_{n}^{(i)}=\operatorname{AvgPool1D}\left(\hat{\mathbf{x}}_{n}^{(i)}\right)}
    \end{equation}
    where $\overline{\mathbf{x}}_{n}^{(i)} \in \mathbb{R}^{d_{\mathbf{x}} \times \frac{T}{k}}$. \equref{eq:avgpool} reduces the memory usage while preserving most of the information in the historical data \cite{zhou2021informer}.
    We empirically show that such down-sampling has little impact on the prediction accuracy (see \secref{sec:analysis:average}).
    \item 
    Then we feed $\overline{\mathbf{x}}_{n}^{(i)}$ into a biGRU to generate the state sequence  $\mathbf{h}_{n}^{(i)}\in \mathbb{R}^{\frac{T}{k} \times d_{h}}$:
    \begin{equation}
        \begin{aligned}
            \medop{\mathbf{h}_{n}^{(i)}= \left\{h_{n,(t)}^{(i)}\right\}_{t=1}^{\frac{T}{k}}}& =\medop{\operatorname{BiGRU}\left(\overline{\mathbf{x}}_{n}^{(i)}\right)}
        \end{aligned}
    \end{equation}
    where $h_{n,(t)}^{(i)} \in \mathbb{R}^{d_h}$ is a hidden state with feature dimension $d_h$.
    Note that $\mathbf{h}_{n}^{(i)}$ can be considered as the distribution characterization for $\overline{\mathbf{x}}_{n}^{(i)}$.
\end{itemize}
The hyper layer explicitly models and characterizes the distributions from the historical data.
Such information is then utilized to generate the weights of the main layers, as explained below.



\subsection{Main Layers}
\label{sec:method:main}
The main layers can adaptively change their parameters conditioned on $\mathbf{h}_{n}^{(i)}$ and $\mathbf{x}^{i}$. 
The  core part of a main layer is a simple GRU cell.
To illustrate its weights generation process, we first present the standard formulation of the GRU cell:

\begin{equation}\label{eq:grucell}
    \begin{aligned}
    \medop{r_{t}}&=\medop{\sigma\left(W_{x r} {x}_{t}^{(i)}+W_{h r} {s}_{t-1}^{(i)}+b_{r}\right)} \\
    \medop{z_{t}}&=\medop{\sigma\left(W_{x z} {x}_{t}^{(i)}+W_{h z} {s}_{t-1}^{(i)}+b_{z}\right)} \\
    \medop{n_{t}}&=\medop{\tanh \left(W_{x n} {x}_{t}^{(i)}+r_{t} \odot \left(W_{h n} \mathbf{s}_{t-1}^{(i)}++b_{n}\right)\right)} \\
    \medop{{s}_{t}^{(i)}}&=\medop{\left(1-z_{t}\right)\odot n_{t}+z_{t} \odot \mathbf{s}_{t-1}^{(i)}}
    \end{aligned}
\end{equation}
where ${x}_{t}^{(i)}\in \mathbb{R}^{d_{\mathbf{x}}}$ denotes input of the GRU cell from  $\mathbf{x}^{(i)}$ at time step $t$, ${s}_{t}^{(i)} \in \mathbb{R}^{d_{s}}$ is the hidden state at $t$, $\sigma$ is sigmoid function, $W_{x r}$, $W_{x z}$, $W_{x n}$, $W_{h r}$,  $W_{h z}$, $W_{h n}$, $b_{r}$, $b_{z}$ and $b_{n}$ are learnable parameters.

The initial hidden state ${s}_{0}^{(i)}$ is computed as follows:
\begin{equation}
    \medop{s_{0}^{(i)}=W_{i n i t} h_{n,\left(\frac{T}{k}\right)}^{(i)} \mid+b_{i n i t}}
\end{equation}
where $h_{n,\left(\frac{T}{k}\right)}^{(i)}$ is the last state in $\mathbf{h}_{n}^{(i)}$; $W_{i n i t}$ and $b_{i n i t}$ are learnable parameters.

The weights of the GRU cell at current time step $t$ are controlled by a vector ${v}_{t}$ that varies dynamically with $t$ :
\begin{equation} \label{eq:weight}
     \medop{{W}_{Ih}={W}_{hv}{v_t},\,{W}_{Ix}={W}_{xv}{v_t},\,{W}_{Ib}={W}_{bv}{v_t}}
\end{equation}
where ${{W}_{hv}}$, ${{W}_{xv}}$ and ${{W}_{bv}}$ are learnable parameters. ${W}_{Ih}\in\mathbb{R}^{d_{Ih}}$, ${W}_{Ix}\in\mathbb{R}^{d_{Ix}}$, and ${W}_{Ib}\in\mathbb{R}^{d_{Ib}}$ are the weights of the GRU cell, ${v}^{t}$ is determined by the previous state $s^{(i)}_{t-1}$ and the outputs of the hyper layers $\mathbf{h}_{n}^{(i)}$, which is computed by the attention mechanism:
\begin{equation} \label{eq:v_attn}
    \begin{aligned}
    \medop{{v}_{t}}= \medop{{W_c}{c}_{t}}, \,
    \medop{{c}_{t}}=\medop{\sum_{p=1}^{\frac{T}{k}} \alpha_{p} {h}_{n,(p)}^{(i)}},\,
    \medop{\alpha_{p}} =\medop{\frac{\exp \left(\operatorname{score}\left({s}^{(i)}_{t-1}, {h}_{n,(p)}^{(i)}\right)\right)}{\sum_{q=1}^{\frac{T}{k}} \exp \left(\mathrm{score}\left({s}^{(i)}_{t-1}, {h}_{n,(q)}^{(i)}\right)\right)}}
    \end{aligned}
\end{equation}
where $\operatorname{score}(\cdot)$ is a score function, according to Luong attention \cite{luong2015effective}.
The score function is given by:
\begin{equation}
   \medop{\operatorname{score}\left({s}^{(i)}_{t-1}, {h}_{n,(p)}^{(i)}\right) = V^{\mathrm{T}}\operatorname{tanh}(W_{s}{s}^{(i)}_{t-1}+W_{h}{h}_{n,(p)}^{(i)}+b_{s})}
\end{equation}
where $V$, $W_{s}$, $W_{h}$, and $b_{s}$ are learnable parameters. 

As mentioned above, all the parameters of the main layer are generated by ${{W}_{Ih}}$, ${{W}_{Ix}}$, and ${{W}_{Ib}}$: 
\begin{equation}
    \begin{aligned}
    \medop{\left(W_{h r}, W_{h z}, W_{h n}\right)} &=\medop{\operatorname{Chunk}\left(W_{I h}\right)} \\
     \medop{\left(W_{x r}, W_{x z}, W_{x n}\right)} &=\medop{\operatorname{Chunk}\left(W_{I x}\right)} \\
    \medop{\left(b_{r}, b_{z}, b_{n}\right)} &=\medop{\operatorname{Chunk}\left(W_{I b}\right)}
    \end{aligned}
\end{equation}
where $W_{h r}$,  $W_{h z}$,  $W_{h n} \in \mathbb{R}^{d_{s} \times d_{s}}$, $d_{s} \times 3d_{s}=d_{Ih}$ and $W_{x r}$,  $W_{x z}$,  $W_{x n}\in \mathbb{R}^{d_{s} \times {d_{\mathbf{x}}}}$ , $d_{\mathbf{s}} \times 3d_{x}=d_{Ix}$  and $b_{r}$, $b_{z}$, $b_{n}\in \mathbb{R}^{d_{s}}$ , $3d_{s}=d_{I b}$. 
$\operatorname{Chunk(\cdot)}$ is the operation to split a tensor into a specific number of equal-sized chunks.
Through \equref{eq:v_attn}, we can adaptively attach different information in $\mathbf{h}_{n}^{(i)}$ to the state of the GRU cell at the current time step $t$ to dynamically adjust the weights of the main layers. 
By integrating \equref{eq:grucell}, \equref{eq:weight} and \equref{eq:v_attn}, we can dynamically generate weights of the main layers.

\subsection{Training}
\label{sec:method:training}

According to \secref{sec:method:meta} and \secref{sec:method:main}, we implement  HyperGRU with a simple prediction layer(\eg sigmoid function) that takes $\hat{\mathbf{x}}_{n}^{(i)}$ and $\mathbf{x}^{(i)}$ as inputs and outputs the prediction:
\begin{equation}\label{eq:predict}
    \medop{\hat{\mathbf{y}}^{(i)}_{n} = \operatorname{HyperGRU}(\mathbf{x}^{(i)}, \hat{\mathbf{x}}_{n}^{(i)})}
\end{equation}
The loss of the $i$-th pair $(\mathbf{x}^{(i)}, \mathbf{y}^{(i)})$ is calculated as follows:
\begin{equation}\label{eq:loss}
    \medop{\mathcal{L}_{\text {pred }}(\theta) =\frac{1}{L} \sum_{n=1}^{L}   \operatorname{criterion}\left(\hat{\mathbf{y}}^{(i)}_{n}, \mathbf{y}^{i}\right)}  
\end{equation}
where $\operatorname{criterion}\left(\cdot,\cdot \right)$ is an objective function, and $\theta$ denotes the learnable parameters of HyperGRU.
Then we perform training using \equref{eq:loss}.

\subsection{Extensions to Other Models}
\label{sec:method:extensions}
As mentioned, \sysname is a model-agnostic framework that can incorporate various time series forecasting models.
We briefly explain how to extend \sysname to other RNNs such as vanilla RNN, LSTM\cite{bib:NC97:Hochreiter}, ConvLSTM\cite{DBLP:conf/nips/ShiCWYWW15}\etal.

Extending \sysname to other RNNs is very straightforward. Take the LSTM as an example, we can use \equref{eq:v_attn} to calculate $v_{t}$, and use $v_{t}$ to generate the weights of memory cell, input gate, forget gate and output gate. The weights of other RNNs can be generated in a similar way.


%% file: body/experiment.tex
\section{Experiments}
\label{sec:exp}

In this section, we evaluate the performance of \sysname on 9 benchmarks and different time series models covering short sequence time series forecasting (\secref{sec:exp:short}), grid-based spatiotemporal data forecasting (\secref{sec:exp:grid}), and graph-based data spatiotemporal forecasting (\secref{sec:exp:graph}). All experiments were 
conducted out on a single Nvidia RTX3090 GPU.


\subsection{Short Sequence Time Series Forecasting}
\label{sec:exp:short}
We test \sysname for short sequence time series forecasting on three datasets from diverse application domains.

\subsubsection{Datasets and Metrics}
\begin{itemize}
    \item \textbf{Air Quality} \cite{zhang2017cautionary}:
    This dataset contains hourly air quality data collected from 12 stations in Beijing, China, from March 2013 to February 2017. 
    We use data from the same four stations (Dongsi, Tiantan, Nongzhanguan, and Dingling) and the same six features (PM2.5, PM10, SO2, NO2, CO, and O3) as \cite{du2021adarnn}. 
    We use the data from 01/03/2013 to 30/06/2016 for training, data from 01/07/2016 to 31/10/2016 for validation, and data from  02/11/016 to 28/02/2017 for testing. 
    The Root Mean Square Error (RMSE) and Mean Absolute Error (MAE) are used as the performance metrics.
    \item \textbf{Electricity Consumption} \cite{electricity}:
    This dataset contains household electric power consumption measurements of 2, 062, 346 valid samples (null values removed).
    The measurements were collected between 16/12/2006 and 26/11/2010.
    Each interval lasts 10 min. 
    We use the data from 16/12/2006 to 24/04/2009 for training, data from 25/02/010 to 26/11/2010 for testing, and the remaining for validation. 
    RMSE is used as the performance metric.
    \item \textbf{Human Activity} \cite{almaslukh2017effective}:
    This dataset contains smartphone sensor data (accelerometer, gyroscope, and magnetometer) collected from 30 volunteers performing six activities.
    We predict the activities of volunteers based on the historical data. 
    We use 7,352 instances for training and 2, 947 instances for testing following \cite{du2021adarnn}.
    We use Accuracy (ACC), Precision (P), Recall (R), and F1 as the performance metrics.
\end{itemize}

\begin{table}[t]
\centering
\footnotesize  
\caption{Datasets for short sequence time series forecasting.}\label{tab:sdataset}
\begin{tabular}{lrrr}
    \toprule
     Datasets&\#Train&\#Valid&\#Test \\
    \midrule
     Air quality &29,232&2,904&2,832 \\
     Electricity consumption &1,235,964&413,280&413,202 \\
     Human activity &6,000&1,352&2,947 \\ 
     \bottomrule
\end{tabular}
\end{table}

\begin{table*}[t]
\centering
\caption{Performance of Air quality and Electricity Consumption.}
\label{tab:air}
\begin{tabular}{l|cccccccc|r}
\hline
\multirow{2}{*}{Methods} & \multicolumn{2}{c}{Dongsi}        & \multicolumn{2}{c}{Tiantan}       & \multicolumn{2}{c}{Nongzhanguan}  & \multicolumn{2}{c|}{Dingling}     & \multicolumn{1}{c}{Electric Power}  \\
                         & RMSE            & MAE             & RMSE            & MAE             & RMSE            & MAE             & RMSE            & MAE             & \multicolumn{1}{c}{RMSE}            \\ \hline
LSTM  & 0.0502          & 0.0403          & 0.0511          & 0.0453          & 0.0341          & 0.0476          & 0.0329          & 0.0339          & 0.092                               \\
GRU  & 0.0510          & 0.0380          & 0.0519          & 0.0475          & 0.0348          & 0.0459          & 0.0330          & 0.0347          & 0.093                               \\
LSTNet  & 0.0544          & 0.0651          & 0.0519          & 0.0651          & 0.0548          & 0.0696          & 0.0599          & 0.0705          & 0.080                               \\
MMD-RNN  & 0.0360          & 0.0267          & 0.0183          & 0.0133          & 0.0267          & 0.0197          & 0.0288          & 0.0168          & 0.082                               \\
DANN-RNN & 0.0356          & 0.0255          & 0.0214          & 0.0157          & 0.0274          & 0.0203          & 0.0291          & 0.0211          & 0.080                               \\
STRIPE & 0.0365          & 0.0216          & 0.0204          & 0.0148           & 0.0248          & 0.0154          & 0.0304          & 0.0139          & 0.086                               \\
ADARNN & 0.0295          & 0.0185          & 0.0164          & 0.0112          & 0.0196          & 0.0122          & 0.0233          & 0.0150          & 0.077                               \\ \hline
HyperLSTM                & \textbf{0.0278} & \textbf{0.0178} & \textbf{0.0155} & \textbf{0.0107} & 0.0189          & 0.0120          & 0.0231          & \textbf{0.0134} & \textbf{0.075}                      \\
HyperGRU               & 0.0285          & 0.0182          & 0.0156          & 0.0111          & \textbf{0.0184} & \textbf{0.0119} & \textbf{0.0229} & 0.0138          & \multicolumn{1}{r}{\textbf{0.075}} \\ \hline
\end{tabular}
\end{table*}

\begin{table}[t]
    \centering
    \footnotesize  
    \caption{Performance of human activity prediction}\label{tab:act}
    \begin{tabular}{lccccc}
    \Xhline{0.5pt}  
    Methods & ACC&P&R&F1&\\
   \Xhline{0.5pt}  
    LightGBM  &84.11 &83.73 &83.63 &84.91 \\
    GRU  &85.68 &85.62 &85.51 &85.46 \\
    MMD-RNN  & 86.39 &86.80 &86.26 &86.38 \\ 
    DANN-RNN  & 85.88 &85.59 &85.62 &85.56 \\
    ADARNN & 88.44 &88.71 &88.59 &88.63 \\
    \Xhline{0.5pt}  
    HyperGRU & \textbf{89.61} &\textbf{89.64}&\textbf{89.45}&\textbf{89.43}\\
    \Xhline{0.5pt}  
    \end{tabular}
\end{table}

\subsubsection{Baselines and Configurations}
We adopt GRU/LSTM as the hyper layers and main layers for our \sysname framework because  they suffice for short sequence prediction in many applications \cite{salinas2020deepar, DBLP:conf/nips/0001YL0020}.
The corresponding models are termed HyperGRU and HyperLSTM.
We compare them with the following baselines:
\begin{itemize}
    \item \textbf{LSTM} : 
    a standard long short memory network.
    \item \textbf{GRU} : 
    a standard gated recurrent unit.
    \item \textbf{LightGBM} : 
    a lightweight and efficient gradient boosting decision tree.
    \item \textbf{MMD-RNN} : 
    a domain generalization method, whose backbone is changed to RNNs and uses maximum mean discrepancy measure as the final output of the RNN layers.
    \item \textbf{DANN-RNN} : 
    the RNN version of a dynamic adversarial adaptation network to dynamically learn domain-invariant representations \cite{du2021adarnn}.
    \item \textbf{LSTNet} : 
    a method combining Convolutional Neural Networks(CNNs) and LSTM for time series forecasting.
    \item \textbf{STRIPE} : 
    an Encoder-Decoder model with a diversification mechanism from a determinantal point process.
    \item \textbf{ADARNN} : 
    the state-of-the-art model to handle distribution shift in time series forecasting.
\end{itemize}
These baselines are chosen for the following reasons.
LSTM and GRU are basic models in our HyperLSTM and HyperGRU. 
LightGBM is a popular and strong traditional machine learning algorithm in Kaggle competitions.
LSTNet and STRIPE are state-of-the-art RNN-based methods without considering the problem of distribution shift. 
MMR-RNN, DANN-RNN, and ADARNN are recent models attempting to solve the distribution shift problem in time series, where ADARNN is state-of-the-art.

For all datasets, we use Adam \cite{bib:ICLR15:kingma} with weight decay regularization as the optimizer.
The dataset-specific configurations are summarized below.
\begin{itemize}
    \item \textbf{Air Quality:} 
    We use L2 loss as the training objective.
    The learning rate is 2e-4 with a weight decay of 0.01, and the batch size is 128. 
    We set sliding window size $T=672$ (28 consecutive days), $k=24$ for 1D average pooling.
    \item \textbf{Electricity Consumption:} 
    We use L2 loss as the training objective and 5e-5 as the learning rate.
    We set sliding window size $T=1008$ (one week), $k=36$ for 1D average pooling, and the batch size is set to 64.
    \item \textbf{Human Activity:} 
    We use cross-entropy as the training objective and the learning rate is 2e-5 with a weight decay of 0.01.
    We set sliding window size $T=128$, $k=36$ for 1D average pooling, and the batch size is set to 256. 
\end{itemize}
We set the prediction step $T_{\mathbf{y}} =1 $ for all benchmarks as in \cite{du2021adarnn}.

\subsubsection{Results}
The left part of \tabref{tab:air} shows the RMSE and MAE of different methods for the air quality prediction tasks.
The right part of \tabref{tab:air} presents the RMSE of predicting the power consumption.
\tabref{tab:act} lists the ACC, P, R, and F1 scores of different methods of activity classification. 
Since LSTNet and STRIPE are built for regression tasks, they are not applicable to this dataset. 
We make the following observations.
\begin{itemize}
    \item 
    HyperLSTM outperforms LSTNet and STRIPE, the methods that ignore distribution shift, by decreasing the RMSE/MAE by 61.4\%/80.6\% and 23.9\% /34.4\% in average.
    This indicates the importance of addressing the distribution shift problem. 
    \item 
    HyperLSTM and HyperGRU outperform MMD-RNN and DANN-RNN on all benchmarks, suggesting that \sysname better handles distribution shift than existing domain adaptation and generalization methods for time series forecasting. 
    \item 
    Compared to ADARNN, one state-of-the-art method for handling distribution shift, HyperLSTM and HyperGRU decrease the RMSE by 3.94\% and 3.82\% averaged on the air quality dataset; and 2.60\% and 2.60\% on the electricity consumption dataset.
    HyperGRU also outperforms ADARNN by increasing ACC by 1.32\% and F1 score by 1.04\% on the human activity dataset.
    These results indicate that \sysname can better characterize the distribution shifts in data. 
\end{itemize}

\subsection{Grid-based Spatiotemporal Forecasting}
\label{sec:exp:grid}
In this subsection, we demonstrate the applicability of \sysname to spatiotemporal data (grid-based).
We are interested in the gains by explicitly accounting for the distribution shifts in spatiotemporal data forecasting.

\subsubsection{Datasets and Metrics.}
We experiment with NYC-Taxi and NYC-Bike, two datasets of trip records collected in New York City (NYC) with grid partitioned regions for traffic prediction. NYC-Taxi contains taxi trip records of NYC from 01/01/2015 to 03/01/2015. 
NYC-Bike contains bike trip records of NYC from 07/01/2016 to 08/29/2016. For both datasets, The first 40 days are used as the training set.
We use the Mean Absolute Percentage Error (MAPE) and RMSE as the performance metrics.

\subsubsection{Baselines and Configurations}
A mainstream strategy to process spatiotemporal data with grid partition is to combine CNNs and RNNs\cite{yao2018deep, yao2019revisiting}.
We choose three representative methods adopting this strategy as the baselines.
\begin{itemize}
    \item \textbf{ConvLSTM} \cite{DBLP:conf/nips/ShiCWYWW15}: 
    an extension of LSTM using convolution operators to build cells.
    \item \textbf{DMVST-Net} \cite{yao2018deep}:
    it combines CNN and LSTM in a joint model that captures the complex relations from both spatial and temporal perspectives.  
     \item \textbf{STDN} \cite{yao2019revisiting}: 
     it designs a periodically shifted attention mechanism to capture long-term dependency and applies LSTM to learn short-term dependency. 
\end{itemize}
We pick these three methods because ConvLSTM is one of the earliest deep learning models for grid-based spatiotemporal data analysis, while DMVST-Net and STDN are very strong baselines for NYC-Taxi and NYC-Bike. 
 

We adapt \sysname to spatiotemporal forecasting as follows.
\begin{itemize}
    \item 
    We adopt ConvLSTM as both the hyper layers and main layers to build the hypernetwork-based model for ConvLSTM, which we term HyperConvLSTM. 
    \item 
    Extensions to STDN and DMVST-Net are more challenging because STDN and DMVST-Net have dedicated components to learn spatial patterns, while \sysname is designed only to learn temporal patterns.
    Therefore, we reuse the corresponding spatial components in STDN and DMVST-Net but replace their temporal components with HyperLSTM.
    The corresponding models are termed STDN w/ HyperLSTM and DMVST-Net w/ HyperLSTM.
\end{itemize}
The region is split into $10\times20$ grids with size $1km \times 1km$.  
The Min-Max normalization is used to convert data value to the $[0,1]$ scale. 
The time interval is set as half-hour.
We set the sliding window size $T=1344$ (four weeks), and $k=48$ for 1D average pooling. L2 loss is used for training.
We optimize all the models by Adam, and the learning rate is set to 2e-4 with a weight decay of 0.01.

\begin{table}[t]
\caption{\sysname adapted for grid-based urban traffic prediction.}\label{tab:nyc} 
\footnotesize
\begin{tabular}{l|l|cccc}

\hline
\multirow{2}{*}{Dataset}  & \multirow{2}{*}{Method} & \multicolumn{2}{c}{pick-up}      & \multicolumn{2}{c}{drop-off}     \\ \cline{3-6} 
                          &                         & RMSE           & MAPE           & RMSE           & MAPE           \\ \hline
\multirow{6}{*}{NYC-Taxi} & ConvLSTM   & 28.13          & 20.50          & 23.67          & 20.70          \\
                          & DMVST-Net & 25.74          & 17.38          & 20.51          & 17.14          \\
                          & STDN                  & 24.10          & 16.30          & 19.05          & 16.25          \\ \cline{2-6} 
                          & HyperConvLSTM           & 26.45          & 19.41          & 22.15          & 29.64          \\
                          & DMVST-Net w/\sysname   & 24.26          & 16.59          & 19.13          & 16.53          \\
                          & STDN w/\sysname        & \textbf{23.26} & \textbf{15.74} & \textbf{18.53} & \textbf{16.26} \\ \hline
\multirow{6}{*}{NYC-Bike} & ConvLSTM   & 10.40          & 25.10          & 9.22           & 23.20          \\
                          & DMVST-Net               & 9.14           & 22.20          & 8.50           & 21.56          \\
                          & STDN                    & 8.85           & 21.84          & 8.15           & 20.87          \\ \cline{2-6} 
                          & HyperConvLSTM           & 9.73           & 24.33          & 8.64           & 21.82          \\
                          & DMVST-Net w/\sysname    & 8.66           & 21.51          & 7.97           & 20.45          \\
                          & STDN w/\sysname        & \textbf{8.13}  & \textbf{21.23} & \textbf{7.62}  & \textbf{19.97} \\ \hline
\end{tabular}
\end{table}

\subsubsection{Results} 
\tabref{tab:nyc} lists the results of traffic flow prediction on NYC-Taxi and NYC-Bike.
We make the following observations.
\begin{itemize}
     \item
    The best performance is achieved with \sysname, indicating the benefit of mitigating distribution shift in traffic prediction.
    \item 
    For the three baselines, the corresponding hypernetwork-based architecture reduces the pick-up RMSE  by 3.49\% to 6.27\% on NYC-Taxi and 5.54\% to 8.13\% on NYC-Bike; and reduces the drop-off RMSE by 2.73\% to 6.73\% on NYC-Taxi and 6.23\% to 6.50\% on NYC-Bike.
    Therefore, our hypernetwork-based framework is easily applicable in grid-based spatiotemporal data forecasting and notably improves prediction accuracy.
\end{itemize}

\subsection{Graph-based Spatiotemporal Forecasting}
\label{sec:exp:graph}
Since graph neural network (GNN) based models have achieved the state-of-the-art performance in urban traffic forecasting\cite{DBLP:conf/nips/0001YL0020,DBLP:conf/aaai/LiZ21}, we are interested in whether \sysname can further improve the performance of GNN-based models.

\subsubsection{Datasets and Metrics.}
We conduct experiments on PeMSD3, PeMSD4, PeMSD7, and PeMSD8, which are graph-based traffic datasets collected by the Caltrans Performance Measurement System \cite{doi:10.3141/1748-12} for traffic flow prediction. 
All the datasets are split with a ratio of 6:2:2 for training, validation, and testing. The traffic flows are aggregated into 5-minute interval. 
\tabref{tab:pemsd} summarizes the detailed information of the datasets.
RMSE, MAE, and MAPE are used as performance metrics.

\begin{table}[t]
\centering
\footnotesize
\caption{Dataset description of PeMSD.}\label{tab:pemsd}
\begin{tabular}{lcc}
    \toprule
     Datasets&\#Detectors& Range\\
    \midrule
     PeMSD3&358&9/1/2018-11/30/2018\\ 
     PeMSD4&307&1/1/2018-2/28/2018\\
     PeMSD7&883&5/1/2017-8/31/2017\\
     PeMSD8&170&7/1/2016-8/31/2016\\
     \bottomrule
\end{tabular}
\end{table}

\subsubsection{Baselines and Configurations}
We use the baselines below.
\begin{itemize}
    \item \textbf{DCRNN}\cite{DBLP:conf/iclr/LiYS018}:  
    an extension of RNN using graph convolutional operators to build cells.
    \item \textbf{STGCN} \cite{DBLP:conf/ijcai/YuYZ18}: 
    it formulates traffic forecasting on graphs and builds the model with GCNs.
     \item \textbf{ASTGCN} \cite{DBLP:conf/aaai/GuoLFSW19}: 
     an attention based spatiotemporal GCN, which can effectively model the dynamic spatiotemporal correlations in traffic data.
     \item \textbf{GraphWaveNet} \cite{DBLP:conf/ijcai/WuPLJZ19}: 
     it can capture the hidden spatial dependency via a novel adaptive dependency matrix.
     \item \textbf{AGCRN} \cite{DBLP:conf/nips/0001YL0020}: 
    an adaptive graph convolutional recurrent network to capture spatial and temporal correlations.
     \item \textbf{STFGNN} \cite{DBLP:conf/aaai/LiZ21}: 
     a spatiotemporal fusion GNN, which can effectively learn hidden spatiotemporal patterns by a fusion operation of various spatial and temporal graphs.
\end{itemize}
The selected GNN-based models include representative (\cite{DBLP:conf/iclr/LiYS018, DBLP:conf/ijcai/YuYZ18}) and state-of-the-art (\cite{DBLP:conf/nips/0001YL0020, DBLP:conf/aaai/LiZ21}) models for traffic forecasting.

We compare these methods with our \sysname using pure RNN-based models (HyperLSTM and HyperGRU) and an extension of \sysname to AGCRN, a GNN-based model.
Specifically, we adopt AG-\\CRN as the hyper layers and main layers to \sysname, and the corresponding model is termed HyperAGCRN. For all four datasets, we use the data of past 12 steps to predict the future 12 steps.
We set sliding window size $T=2016$ (one week), and $k=24$ for 1D average pooling.
We optimize all the models by Adam with batch size 128.
The learning rate is set to 3e-4 with a weight decay of 0.01.

\begin{table*}[t]
\centering
\footnotesize
\caption{\sysname adapted for graph-based traffic prediction.}\label{tab:pems}
\begin{tabular}{l|ccc|ccc|ccc|ccc}
\hline
\multirow{2}{*}{Methods} & \multicolumn{3}{c|}{PeMSD3}     & \multicolumn{3}{c|}{PeMSD4}     & \multicolumn{3}{c|}{PeMSD7}     & \multicolumn{3}{c}{PeMSD8}      \\ \cline{2-13} 
& RMSE & MAE&MAPE  & RMSE&MAE&MAPE    &RMSE& MAE&MAPE    &RMSE&MAE&MAPE 
\\ \hline
LSTM 
& 35.11 & 21.33&21.33  & 39.59 & 25.14&20.33  & 42.84& 29.98&15.33  & 32.06& 22.20&15.32          \\
GRU  
& 36.23& 21.47&21.06 & 38.87 & 25.36&20.49  & 41.93&30.14&15.71  & 32.24&23.03&15.62    \\
DCRNN 
& 30.31 & 18.18&18.91  & 38.12& 24.70&17.12  & 38.58 & 28.30&11.66  & 27.83 & 17.86&11.45          \\
STGCN 
& 30.12& 17.49&17.15   & 35.55& 22.70&14.59   & 38.78& 25.38&11.08   & 27.83 & 18.02&11.40   \\
ASTGCN 
& 29.66 & 17.69&19.40    & 35.22& 22.93&16.56  & 42.57& 28.05&13.92     & 28.16 & 18.61&13.08  \\
GraphWaveNet 
& 32.94& 19.85&19.31   & 39.70& 25.45&17.29  & 42.78& 26.85&12.12   & 31.05& 19.13&12.68  \\
AGCRN 
& -  & - &-        & 32.30& 19.83&12.97         &- & -& -     & 25.22& 15.95&10.09          \\ 
STFGNN 
&26.28 & 16.77&16.30  & 32.51 & 20.48&16.77  &36.60 & 23.46&9.21     & 26.25& 16.94&10.60 \\
\hline
HyperLSTM
& 32.36& 19.21&19.04  & 37.19& 23.65& 18.46  & 38.67 & 27.18& 13.17  & 28.12 & 19.20&13.68  \\
HyperGRU 
& 32.18 & 19.42&18.79  & 36.94& 23.47&19.14     & 38.75& 27.31&12.73 & 27.84& 19.33& 14.23  \\
HyperAGCRN             
& \textbf{25.68} & \textbf{15.85} &\textbf{14.37}
& \textbf{29.93} & \textbf{18.89} &\textbf{12.11}
& \textbf{34.37} & \textbf{21.64} &\textbf{8.89}
& \textbf{24.37} & \textbf{15.26} &\textbf{9.64}
\\ \hline
\end{tabular}
\end{table*}

\subsubsection{Results} 
\tabref{tab:pems} summarizes the average results on graph-based spatiotemporal forecasting.
We have the following observations.
\begin{itemize}
    \item 
    GNN-based models outperform LSTM and GRU since GNN-based models take advantage of modeling spatial patterns while LSTM and GRU only model the temporal patterns.
    \item
    Our HyperLSTM and HyperGRU, despite modeling only the temporal patterns, outperform multiple GNN-based models.
    For example, HyperLSTM and HyperLSTM have better RMSE, MAE, and MAPE than GraphWaveNet on PeMSD3, PeMSD4, and PeMSD7.
    The reason might be that the temporal distribution shift may be a more important bottleneck than the spatial patterns in the accuracy of traffic prediction.
    \item 
    HyperAGCRN achieves the best performance across all datasets. 
    This demonstrates that \sysname can improve the performance of GNN-based models.
\end{itemize}

%% file: body/analysis.tex
\section{Ablation Studies}
\label{sec:analysis}

\subsection{Visual Analysis of Distribution Shift}
\subsubsection{Setups}
In this study, we demonstrate the problem of distribution shift in time series via visual analysis of the Air Quality dataset.
For each station, we randomly sample one instance $\mathbf{x}^{(i)}$ with its corresponding historical dataset $S= \left\{\hat{\mathbf{x}}_{n}^{(i)}\right\}_{n=1}^{L}$, where $S$ contains 128 sequences, \ie $L=128$. 
We use t-SNE \cite{2008Visualizing} to visualize the learned distributions (via the hyper layers) of all sequences in $S$ and their original distributions. 

\subsubsection{Results}
\begin{figure}[t]
    \centering
    \subfloat[]{
        \includegraphics[width=0.24\linewidth]{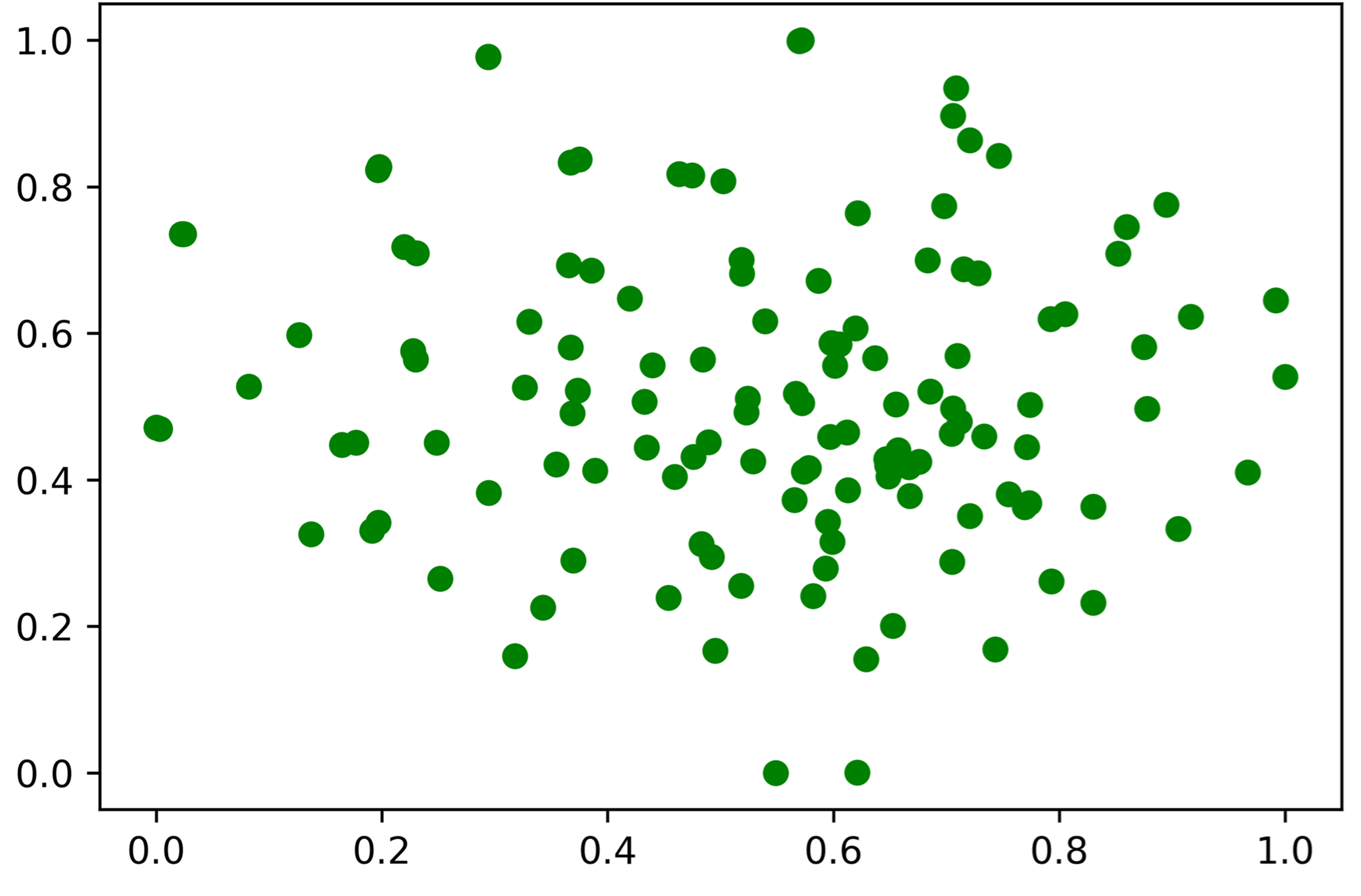}}
    \subfloat[]{
        \includegraphics[width=0.24\linewidth]{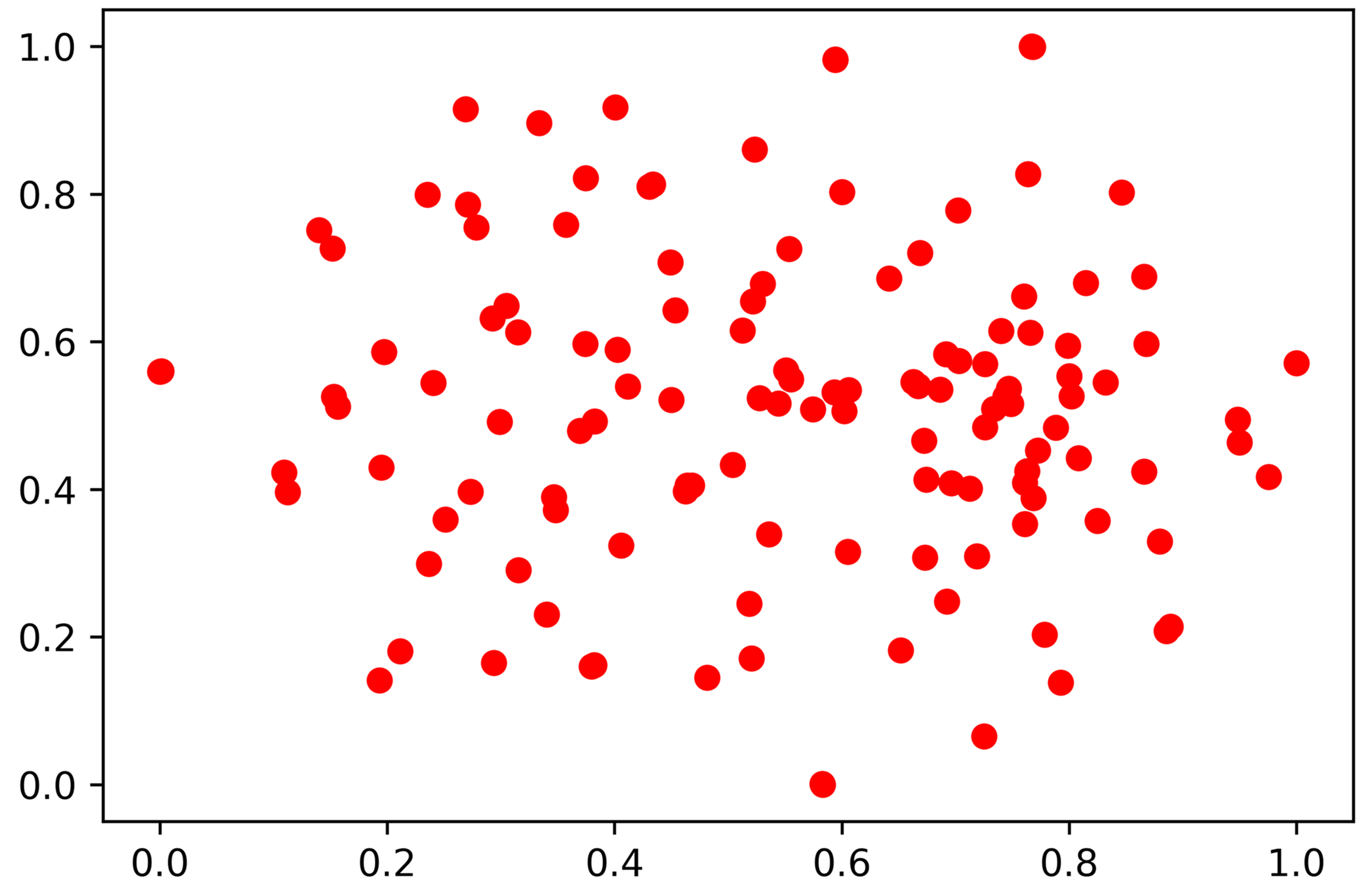}}
    \subfloat[]{
       \includegraphics[width=0.24\linewidth]{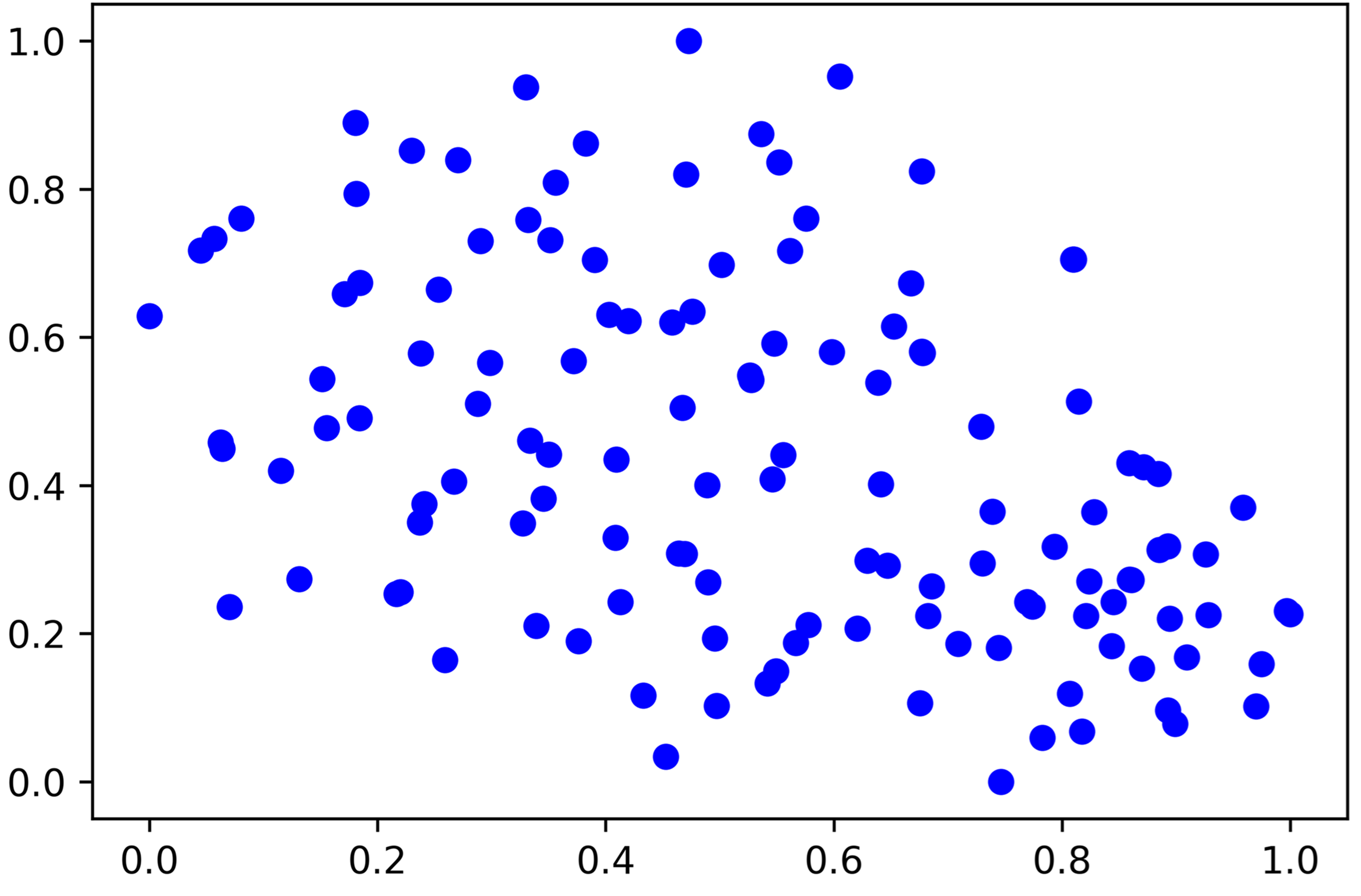}}
     \subfloat[]{
       \includegraphics[width=0.24\linewidth]{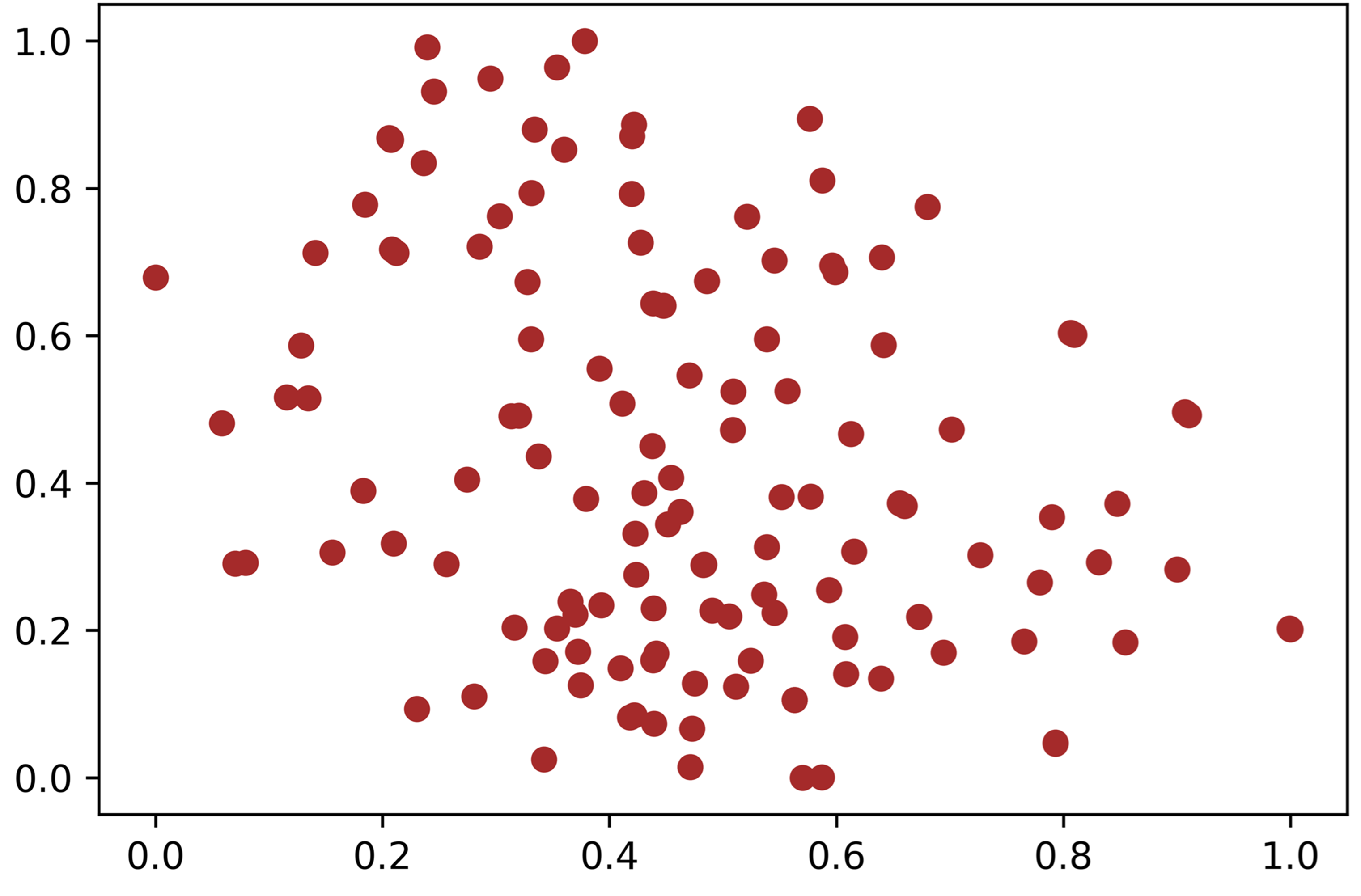}}
    \\
    \subfloat[]{
        \includegraphics[width=0.24\linewidth]{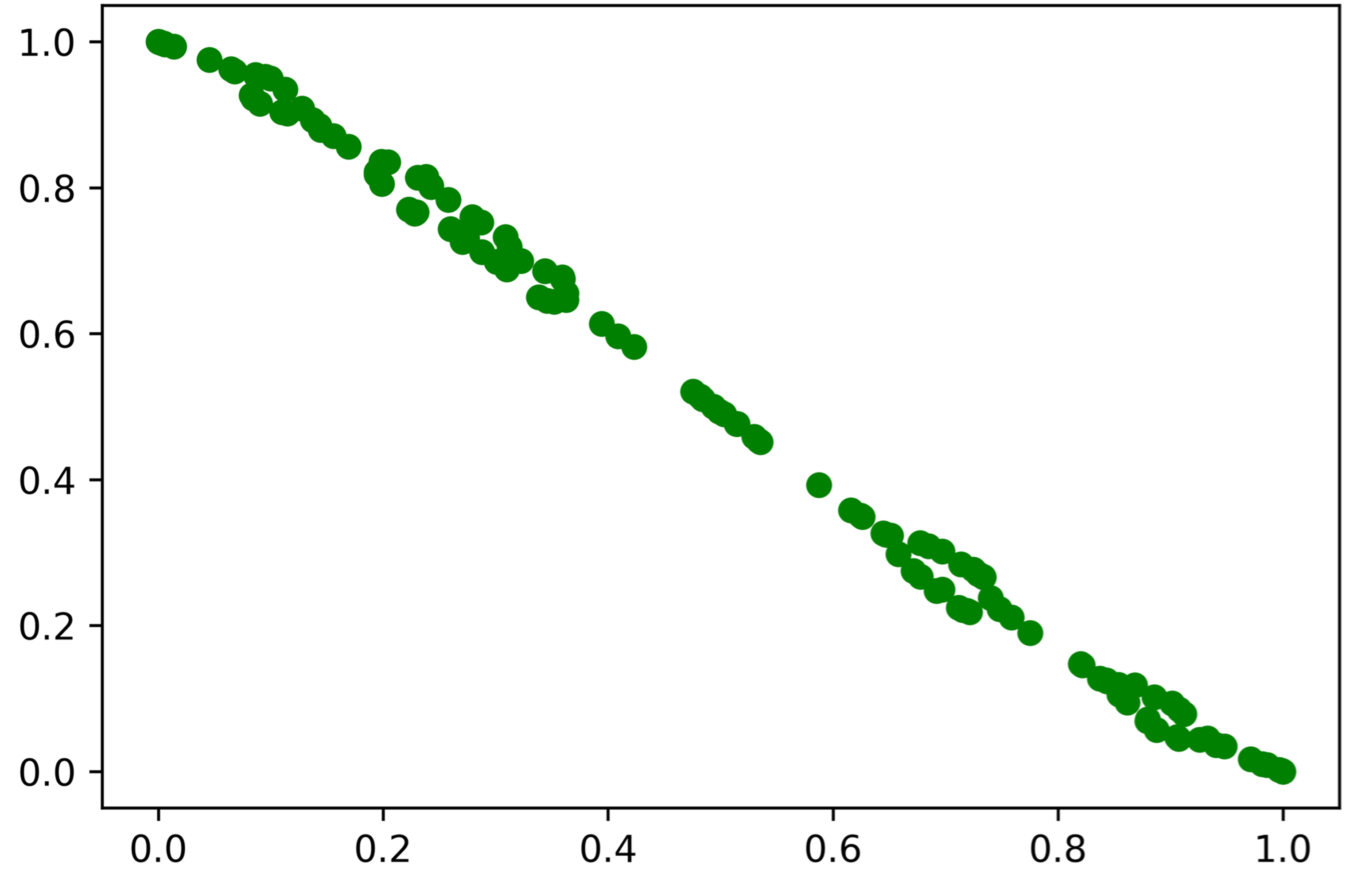}}
    \subfloat[]{
        \includegraphics[width=0.24\linewidth]{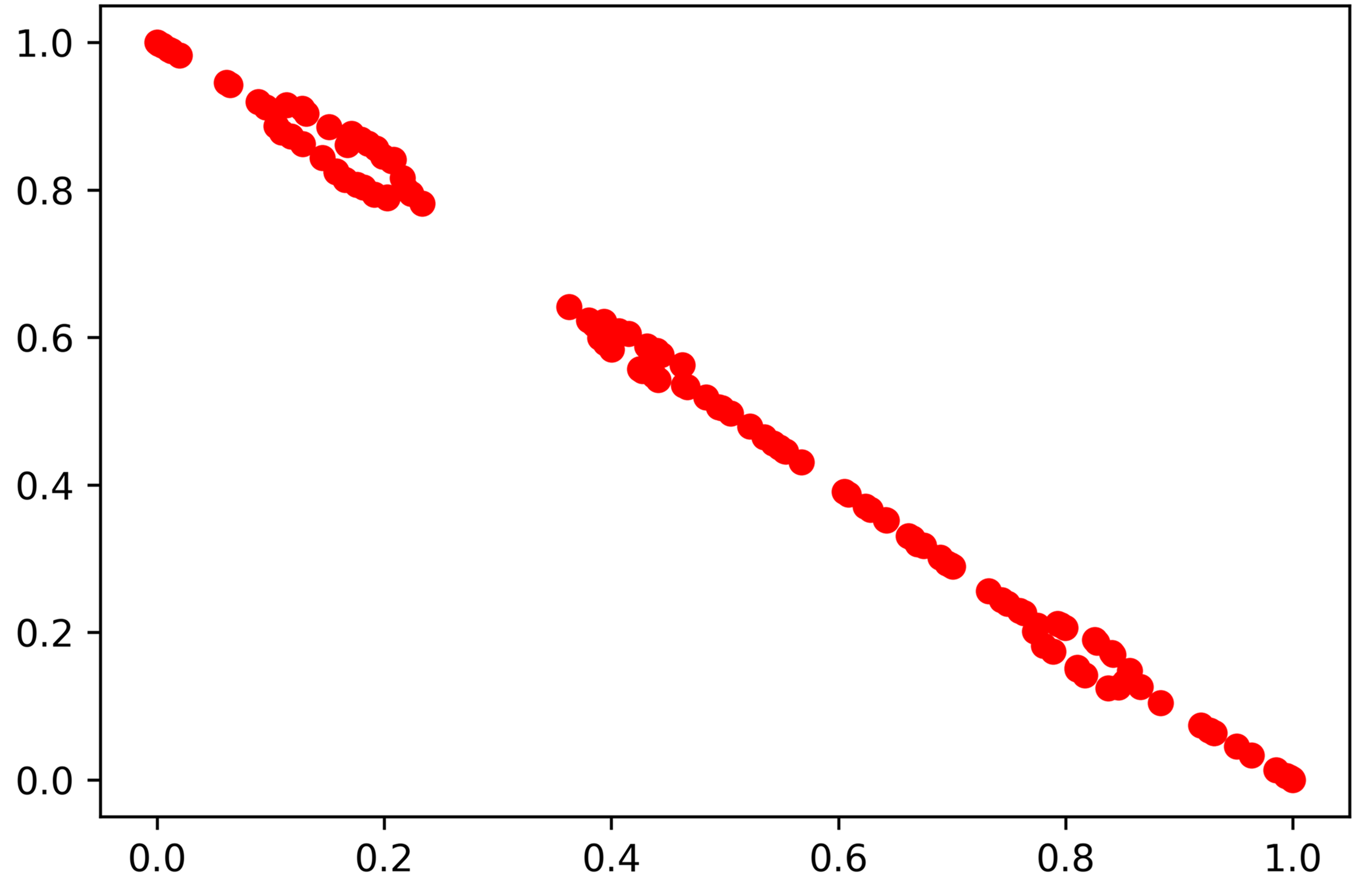}}
    \subfloat[]{
        \includegraphics[width=0.24\linewidth]{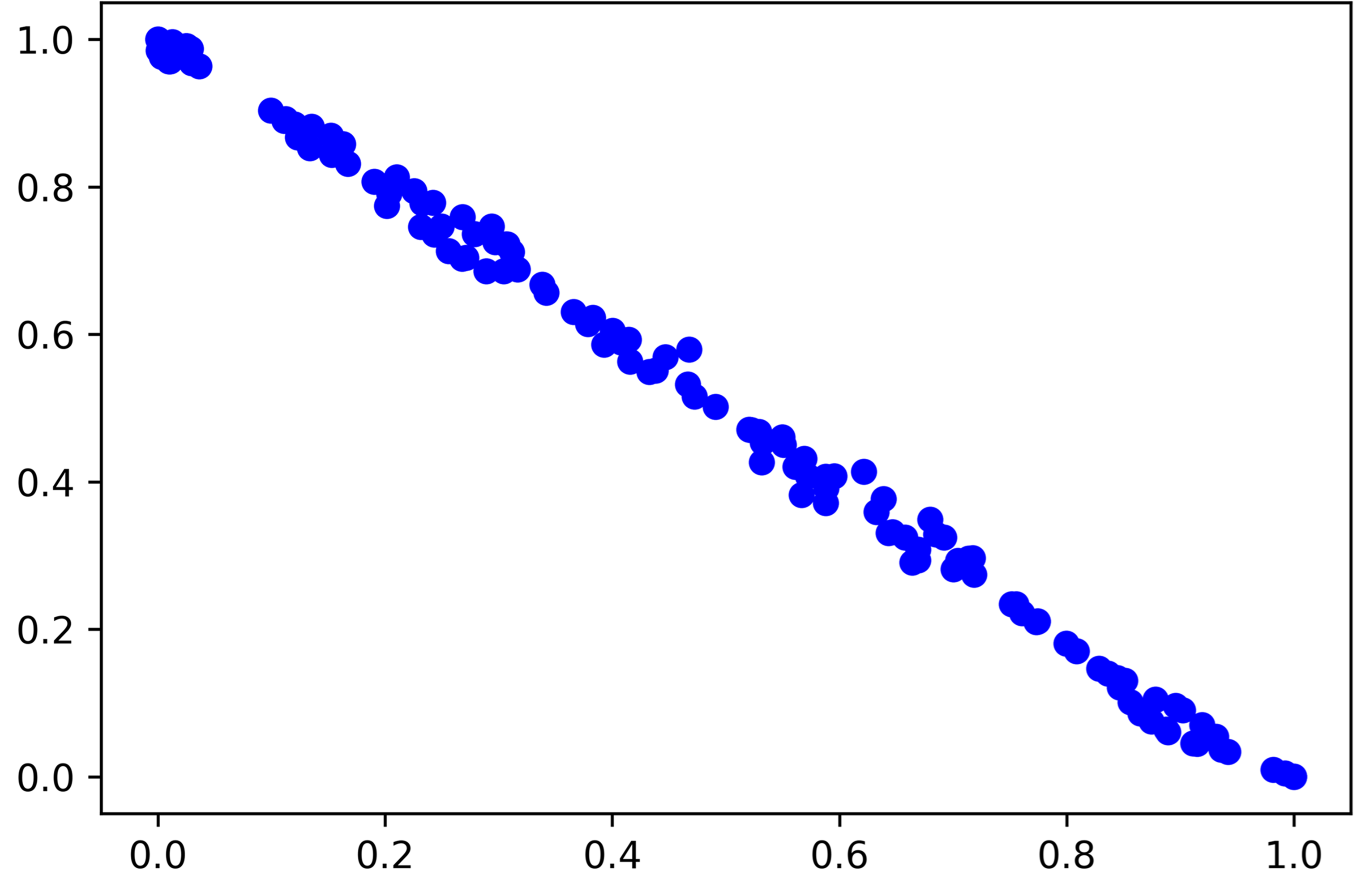}}
    \subfloat[]{
        \includegraphics[width=0.24\linewidth]{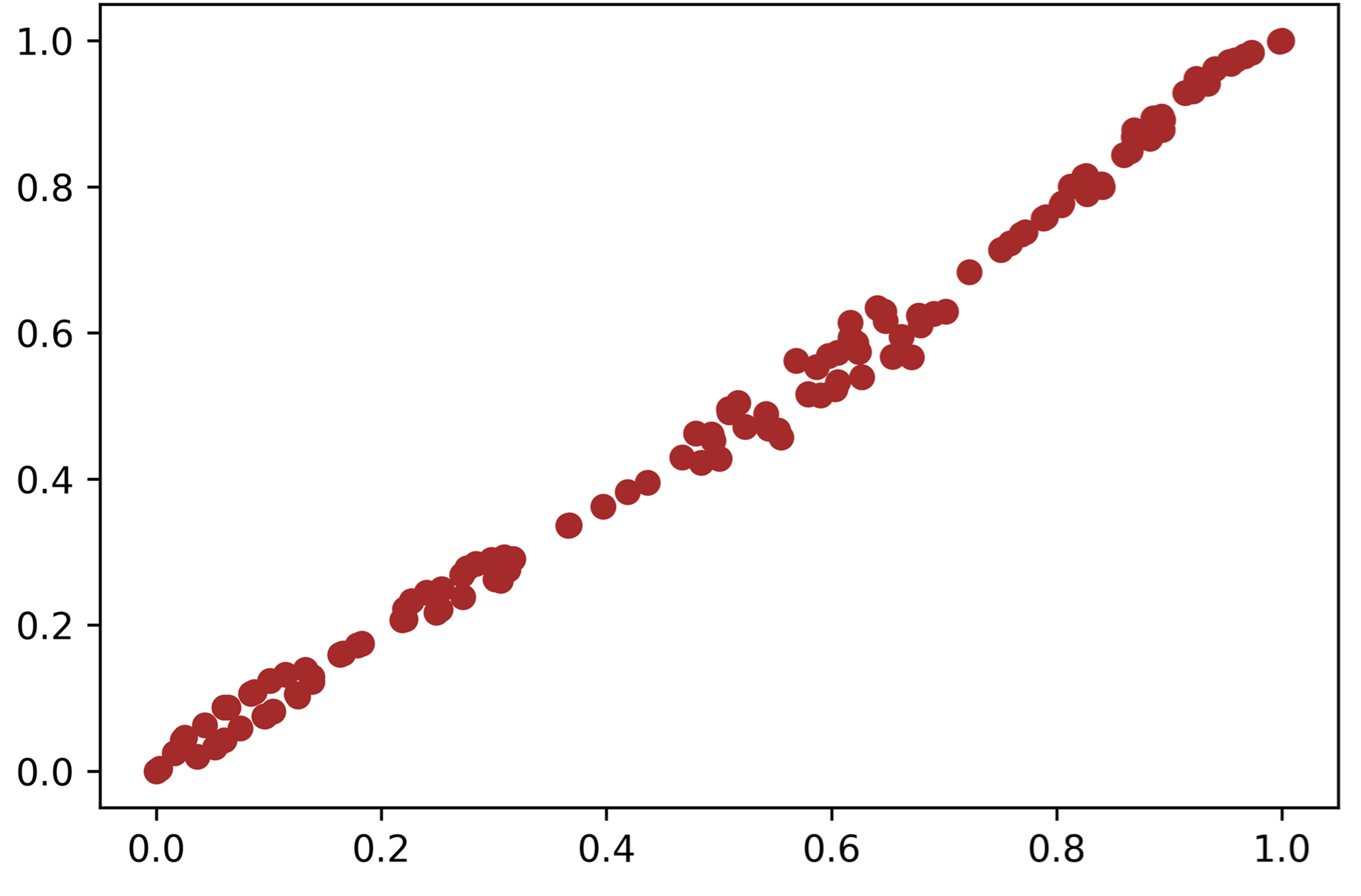}}
    \caption{Visualization of distribution shifts in the Air Quality dataset. The upperparts (a, b, c, d) present the original distribution visualization of four stations (Dongsi, Tiantan, Nongzhangguan, Dingling). The lower parts (e, f, g, h) present the learned distribution visualization of four stations (Dongsi, Tiantan, Nongzhangguan, Dingling). One scatter presents a sequence in $S$.}
   \label{fig:vis}
\end{figure}

From \figref{fig:vis}, we observe that the original distributions are random while the learned distributions can be approximately fitted by a linear model, indicating that common knowledge among all historical sequences has been learned by \sysname.


\subsection{Impact of Sliding Window Length}
\subsubsection{Setups}
The length of the sliding window $T$ is an important hyper-parameter in \sysname.
We experiment with different lengths using HyperGRU on the datasets of Air Quality, NYC-Taxi and NYC-Bike.
As mentioned in \secref{sec:method:meta}, the actual input length of the hyper layers is $\frac{T}{k}$. 
Let $T_{k} = \frac{T}{k}$,
to observe how sensitive the model is to $T_{K}$, we vary $T_{k}$ from 2 to 256. Length 2 is the shortest length of a sequence, and Length 256 is challenging for optimization of GRU. 

\subsubsection{Results}
\begin{figure} [t]
	\centering
	\subfloat[Air Quality]{\label{fig:win:a}
		\includegraphics[scale=0.25]{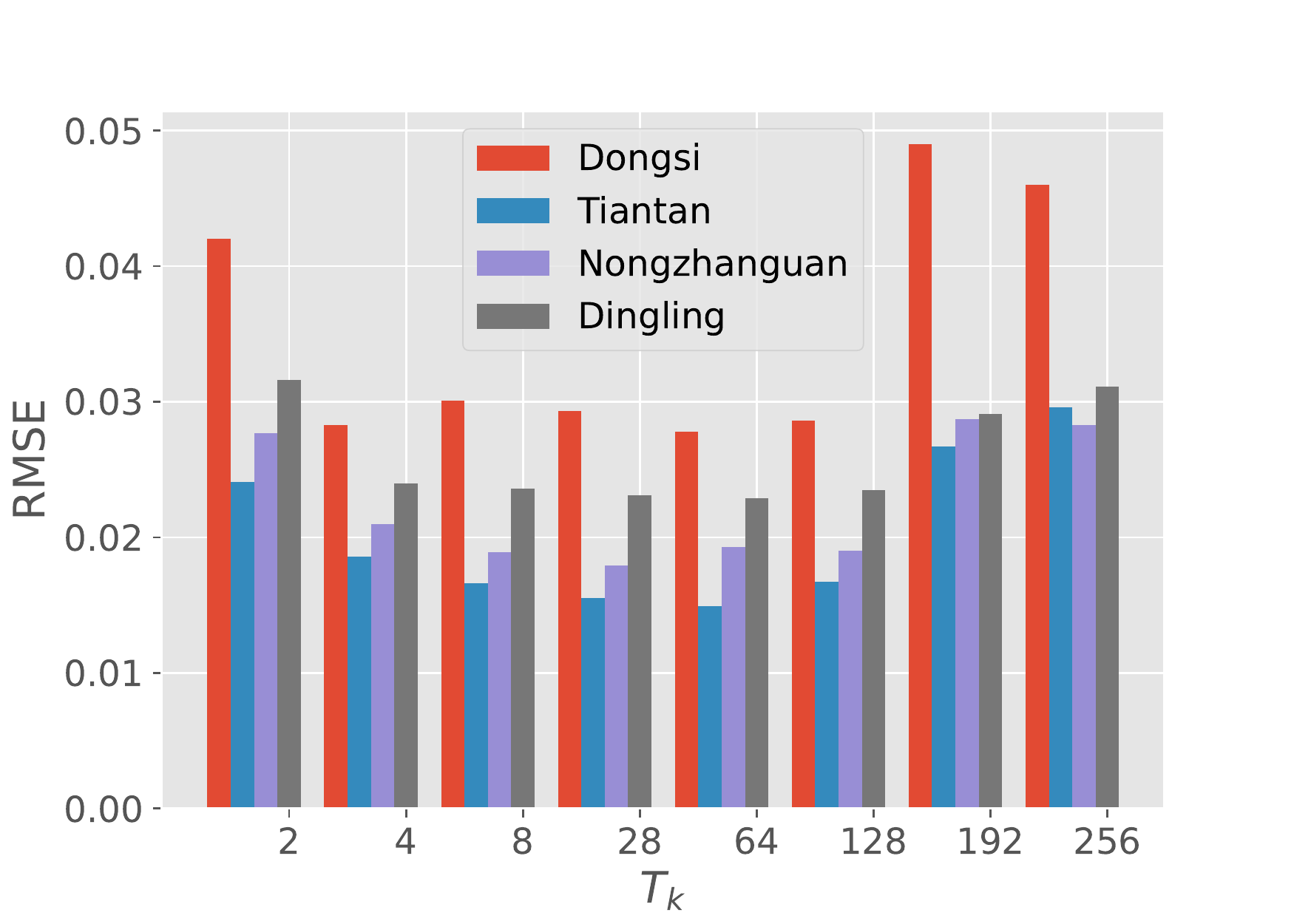}}
	\subfloat[NYC]{\label{fig:win:b}
		\includegraphics[scale=0.25]{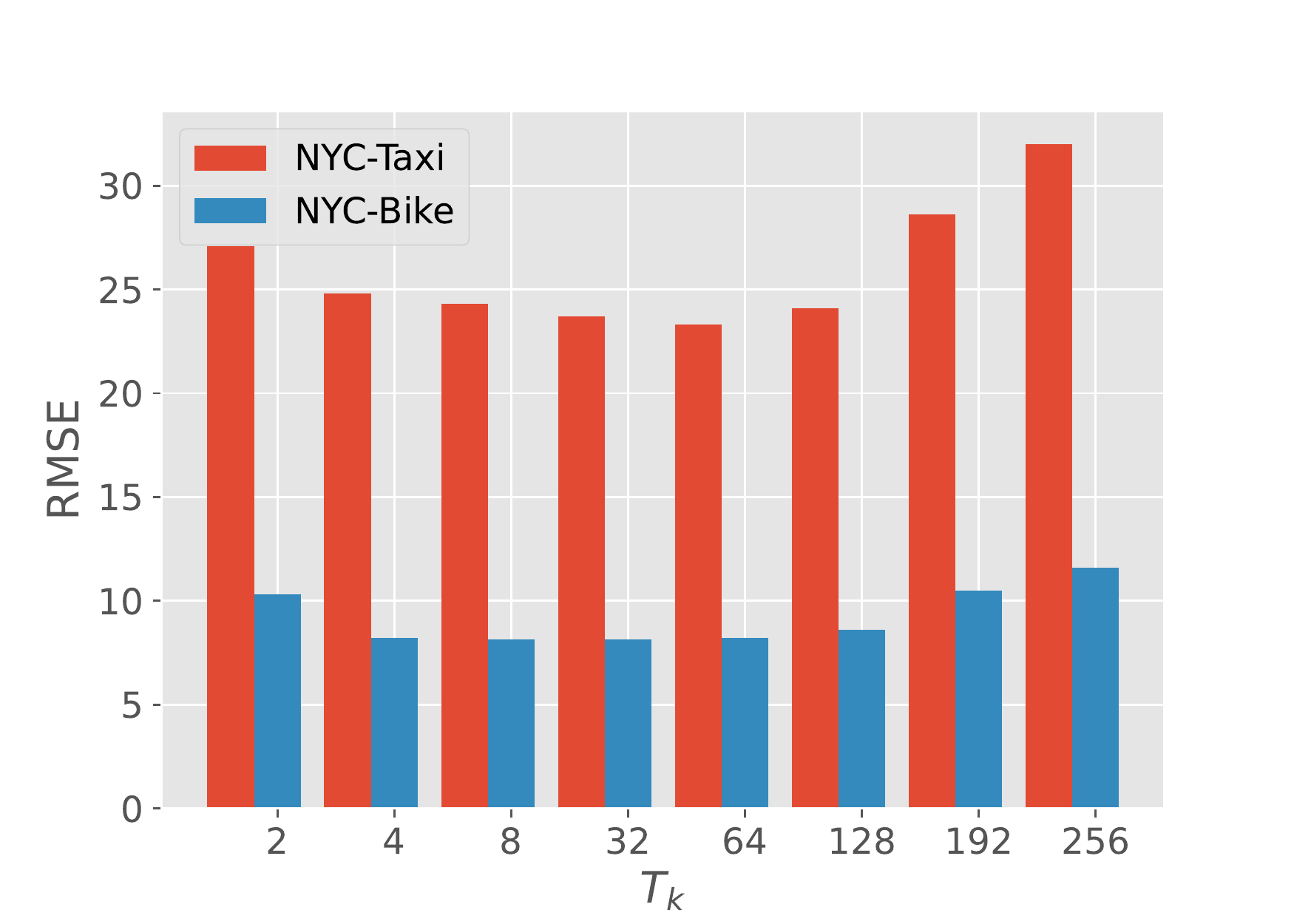}}
	\caption{Impact of sliding window size $T$ on the forecasting accuracy.}
	\label{fig:win} 
\end{figure}

\figref{fig:win} plots the performance variations of HyperGRU on Air-quality, NYC-Taxi and NYC-Bike as the sliding window length increases.
We observe high error (in RMSE) when $T_{k}=2$. 
One potential reason may be that it is difficult to learn complex feature representations of the distribution shifts given very short sequences.
The RMSE only experiences a small variance when $T_{k}$ increases from 4 to 128.
When $T_{k}\geq 192$, the RMSE starts to grow again. One potential reason is that when considering a very long length, the training becomes harder.
This results  show that \sysname is insensitive to sliding window length, except in extreme cases($T_{k}=2$ or $T_{k}\geq 192$).

\subsection{Impact of Prediction Horizon}
\subsubsection{Setups}
Multi-step prediction capability is crucial for traffic flow forecasting.
We analyze the performance of HyperGRU for multi-step prediction on PeMSD4 and PeMSD8 datasets. 

\subsubsection{Results}
\begin{figure} [t]
	\centering
	\subfloat[PeMSD4]{\label{fig:PEMSD:a}
		\includegraphics[scale=0.23]{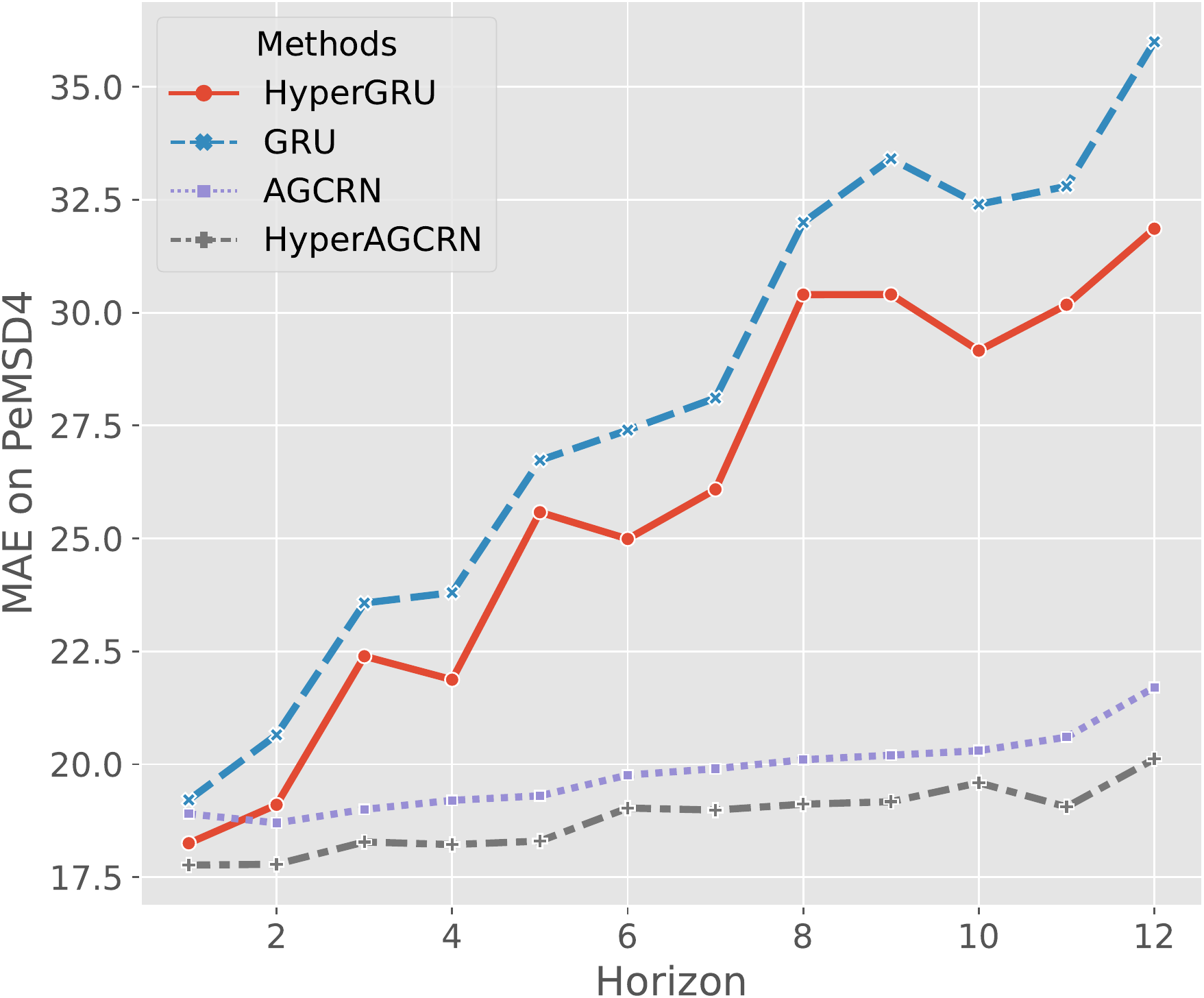}}
	\subfloat[PeMSD8]{\label{fig:PEMSD:b}
		\includegraphics[scale=0.231]{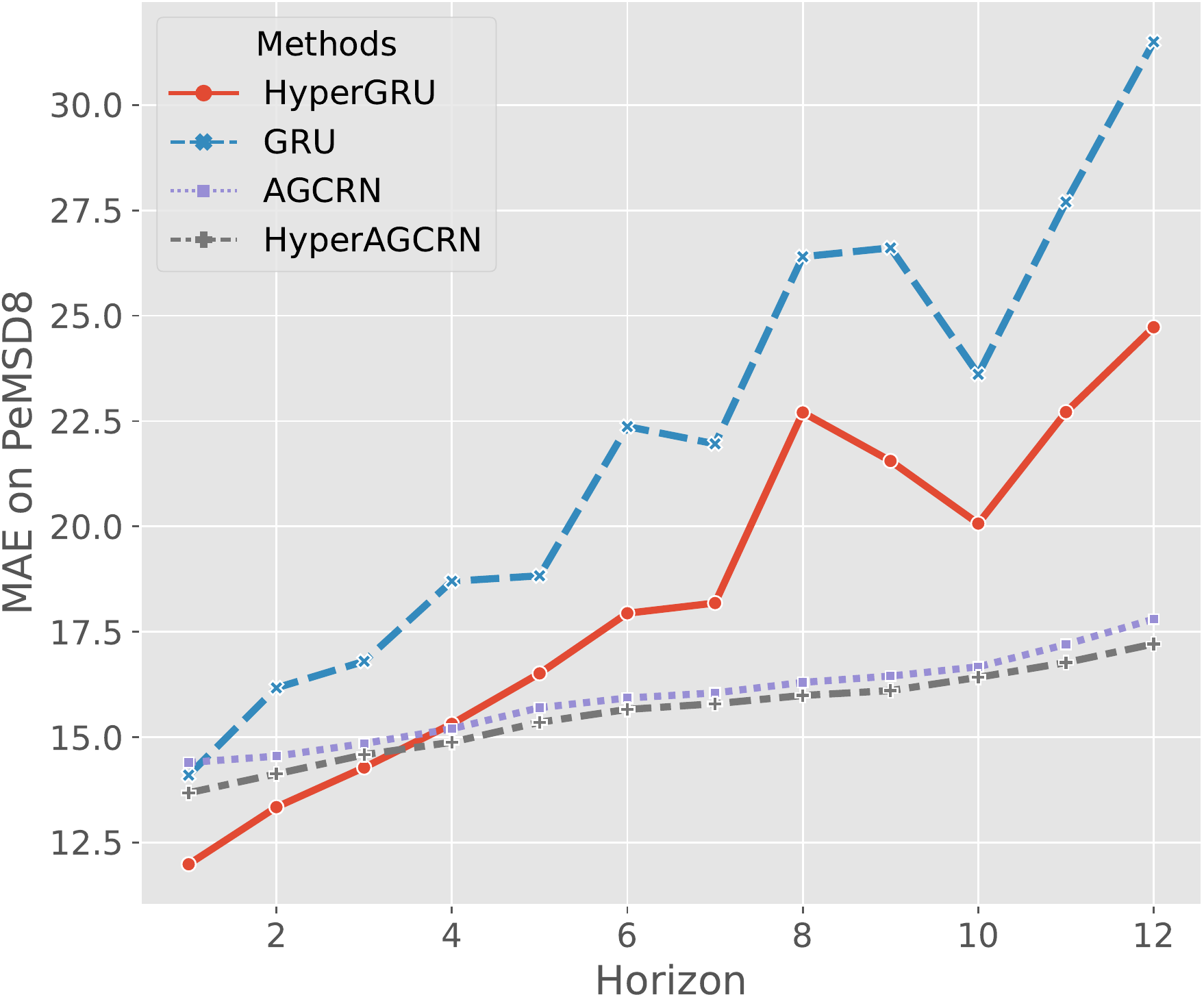}}
	\caption{Multi-step prediction performance on PeMSD.}
	\label{fig:multi}
\end{figure}

\figref{fig:multi} shows the prediction performance of various methods as the horizon increases. 
Overall, the errors (in MAE) grow with the forecast step. 
However, for both datasets, the errors of AGCRN and HyperAGCRN increase much more slowly than GRU and HyperGRU. 
\sysname significantly improves the performance of GRU and HyperAGCRN on multi-step prediction, while our HyperAGCRN achieves the best results for all horizons. 
These results indicate that mitigating the distribution shift problem using \sysname is effective in mining the dynamic patterns of spatiotemporal data.


%% file: body/conclusion.tex
\section{Conclusion}
\label{sec:conclusion}
In this paper, we investigate the distribution shift problem in the time series forecasting problem, which causes discrepancies between the distributions of the training and the testing data.
To this end, we propose \sysname, a novel hypernetwork-based framework which applies for time series forecasting under distribution shift.
Specifically, \sysname exploits the hyper layers to learn the best characterization of the distribution shifts, generating the model parameters for the main layers to make accurate predictions. 
Moreover, \sysname is implemented as an extensible framework that can incorporate diverse time series forecasting models such as RNNs and Transformers. 
Extensive experiments show that \sysname outperforms other state-of-the-art methods on 9 benchmarks.